\documentclass{article}

\PassOptionsToPackage{numbers, compress}{natbib}

\usepackage[preprint]{neurips_2022}

\usepackage[utf8]{inputenc} 
\usepackage[T1]{fontenc}    
\usepackage{hyperref}       
\usepackage{url}            
\usepackage{booktabs}       
\usepackage{amsfonts}       
\usepackage{nicefrac}       
\usepackage{microtype}      
\usepackage{xcolor}         

\usepackage{graphicx}
\usepackage{amsmath}
\usepackage{amssymb}
\usepackage{mathrsfs}
\usepackage{multirow}
\usepackage{subcaption}

\title{DNN-Compressed Domain Visual Recognition with Feature Adaptation}

\author{%
  Yingpeng Deng \quad\quad\quad\quad\quad\quad Lina J. Karam\\
  Image, Video and Usability Lab, School of ECEE, Arizona State University, USA\\
  \texttt{\{ypdeng,karam\}@asu.edu} \\
}

\begin{document}

\maketitle

\begin{abstract}
  Learning-based image compression was shown to achieve a competitive performance with state-of-the-art transform-based codecs. This motivated the development of new learning-based visual compression standards such as JPEG-AI. Of particular interest to these emerging standards is the development of learning-based image compression systems targeting both humans and machines. This paper is concerned with learning-based compression schemes whose compressed-domain representations can be utilized to perform visual processing and computer vision tasks directly in the compressed domain. In our work, we adopt a learning-based compressed-domain classification framework for performing visual recognition using the compressed-domain latent representation at varying bit-rates. We propose a novel feature adaptation module integrating a lightweight attention model to adaptively emphasize and enhance the key features within the extracted channel-wise information. Also, we design an adaptation training strategy to utilize the pretrained pixel-domain weights. For comparison, in addition to the performance results that are obtained using our proposed latent-based compressed-domain method, we also present performance results using compressed but fully decoded images in the pixel domain as well as original uncompressed images. The obtained performance results show that our proposed compressed-domain classification model can distinctly outperform the existing compressed-domain classification models, and that it can also yield similar accuracy results with a much higher computational efficiency as compared to the pixel-domain models that are trained using fully decoded images.
\end{abstract}

\section{Introduction}
\label{sec:intro}

Learning-based image coding algorithms have shown a competitive compression efficiency as compared to the conventional compression methods, by using advanced deep learning techniques. Specifically, when compared to the JPEG/JPEG 2000 standards, learning-based codecs can produce a higher visual quality in terms of PSNR (peak signal-to-noise ratio) and SSIM (structural similarity)~\cite{wang2004image} / MS-SSIM (multi-scale structural similarity)~\cite{wang2003multiscale} for some target compression bitrates~\cite{balle2018variational, Theis2017a}.

\begin{figure}[!tb]
	\centering
	\includegraphics[width=1\textwidth]{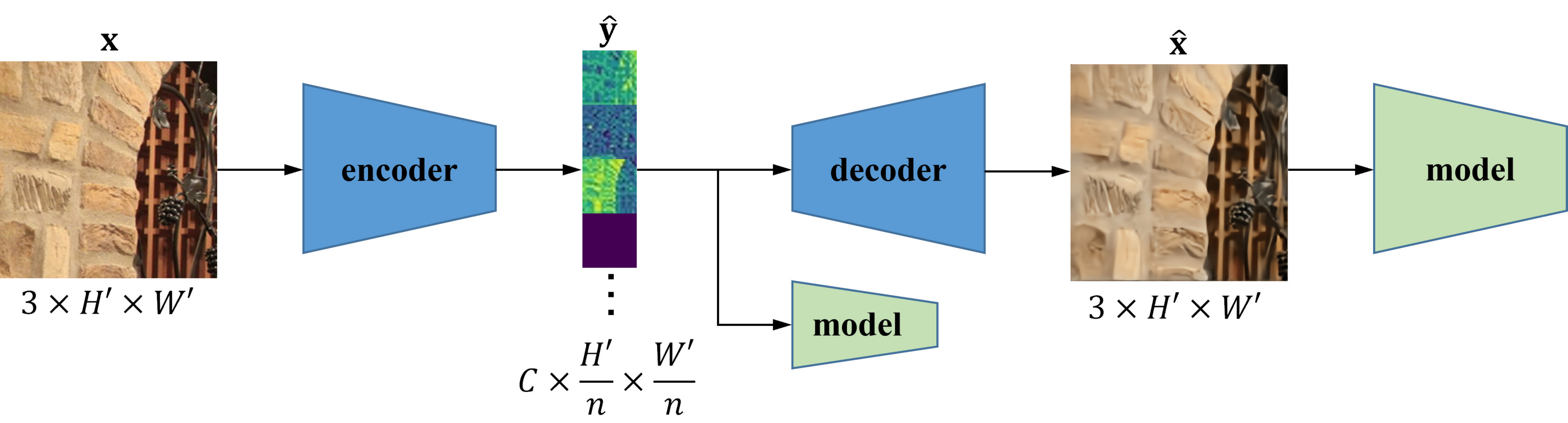}
	\caption{Block diagram of compressed-domain and pixel-domain classifications.}
	\label{fig:flow1}
\end{figure}

Furthermore, without fully decoding the images, learning-based codecs can be adapted to support image processing and computer vision tasks in the compressed-domain. This latter feature was incorporated as part of the emerging JPEG-AI standard~\cite{jpegai_use}. Figure~\ref{fig:flow1} shows an example for a compressed-domain classification task. Given an original input image $\mathbf{x}$, a learning-based codec can encode it and quantize it to form a compressed-domain latent representation $\mathbf{\hat{y}}$ (compressed representation). This compressed representation $\mathbf{\hat{y}}$ can be losslessly transformed into a bitstream via entropy encoding and then recovered back to the compressed-domain through an entropy decoding process. The decoded image $\mathbf{\hat{x}}$ can be reconstructed by providing the compressed representation $\mathbf{\hat{y}}$ as input to the learning-based decoder, and can subsequently be used for training/testing a deep neural network (DNN)-based classifier operating in the pixel domain. Alternatively, without the decoder, the DNN-based classification can be performed directly in the compressed domain when targeting a machine given that the compressed representation $\mathbf{\hat{y}}$ may contain useful information to perform image processing and computer vision tasks in the compressed domain~\cite{jpegai_eval}.

One application of interest is compressed-domain classification. Recently, Torfason~\textit{et al.}~\cite{torfason2018towards} demonstrated that two computer vision tasks, image classification and semantic segmentation, can achieve comparable performance using compressed-domain representations as compared to the tasks performed using pixel-domain decompressed images. While natural image classification can be a common benchmark case, in this work we also consider texture recognition, which plays a fundamental role supporting various classification applications. In the real world, textures are ubiquitous in both natural objects (e.g., flowers, grass, leaves, ripples of waters, rocks) and man-made objects (e.g., floor tiles, printed fabrics, bricks on wall, carpets). They form essential features with visually distinguishable properties that help with segmentation, recognition, and classification. Due to the importance of textures for image and object classification applications and the amazing progress of DNNs, several DNN-based approaches were proposed for texture analysis and recognition~\cite{cimpoi2015deep,dai2017fason,zhang2017deep,zhai2019deep,chen2021deep,xu2021encoding}. Moreover, convolutional layers in deep neural networks act like local feature extractors, which can capture useful local texture information~\cite{zhang2017deep}. Geirhos \textit{et al.}~\cite{geirhos2018imagenet} have shown that ImageNet-trained DNNs rely on textures rather than shapes to recognize images/objects, which indicates that textures play a pivotal role in the success of the DNN models for object recognition.

\begin{figure}[!tb]
	\centering
	\begin{minipage}{1\textwidth}
		\includegraphics[width=1\textwidth]{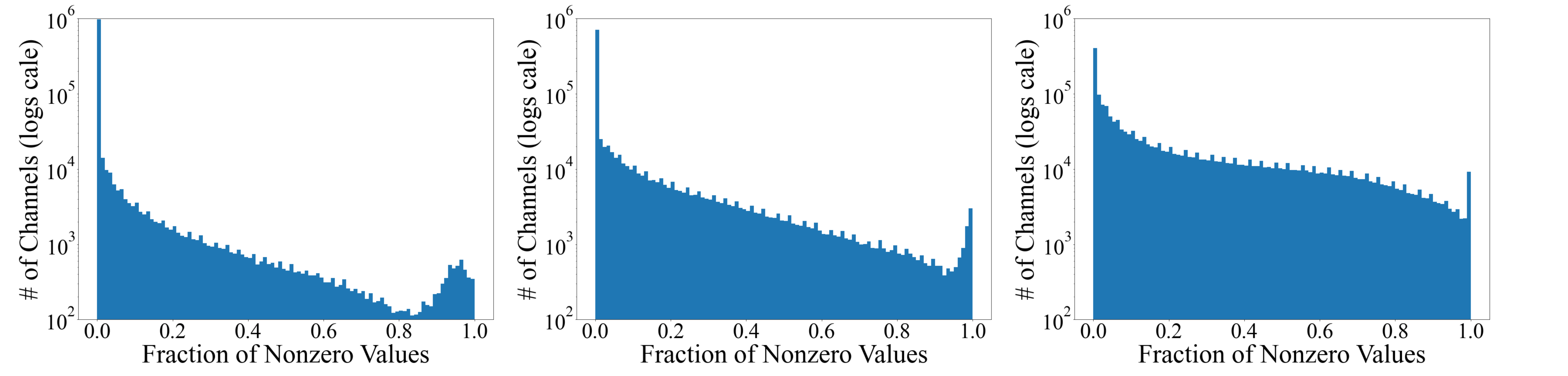}
	\end{minipage}\\
	\small
	\begin{minipage}{0.28\textwidth}
		\centering
		HyperMS-SSIM-1
	\end{minipage}
	\begin{minipage}{0.28\textwidth}
		\centering
		HyperMS-SSIM-4
	\end{minipage}
	\begin{minipage}{0.28\textwidth}
		\centering
		HyperMS-SSIM-8
	\end{minipage}
	\caption{The histograms of compressed-domain representations encoded by using HyperMS-SSIM compression models~\cite{balle2018variational} at different bit rates using the MINC dataset~\cite{bell15minc}. The x-axis corresponds to the fraction of nonzero-value pixels within each single latent representation channel.}
	\label{fig:distr}
\end{figure}

In this work, we adopt a learning-based compressed-domain classification framework~\cite{jpegai_eval,torfason2018towards} for compressed-domain visual recognition. We first evaluate the effects of decoded images on CNN-based recognition model performance and implement the CNN-based compressed-domain recognition using an architecture similar to the one in~\cite{torfason2018towards}. However, the obtained performance results show a clear performance drop for the latter. Moreover, compressed-domain representations are highly sparse at low to mid compression bitrates. According to Figure~\ref{fig:distr}, it can be observed that a high portion of compressed-domain channels contains no or a small number of non-zero values. This majority of zero values may cause zero or little gradient update via backpropagation during the training phase of the compressed-domain DNN models, which results in a significantly reduced performance in terms of classification accuracy. In order to improve the compressed-domain classification performance, we propose a novel feature adaptation (FA) module that can be integrated within the DNN-based classifier for compressed-domain visual recognition. The proposed feature adaptation module consists of two parts - channel-wise attention unit (CAU) and feature enhancement unit (FEU). Specifically, the CAU aims to learn the channel-wise affine transform vectors for selecting and realigning compressed-domain channels, while the FEU aims to enhance the useful features within the selected channels by means of learned cross-channel convolutional layers. Furthermore, we design a training strategy to utilize the pretrained weights of the pixel-domain models. For this purpose, we freeze the adopted weights and only update the parameters of the inserted FA module; then we unfreeze and finetune the whole network for compressed-domain recognition. The obtained performance results show that our proposed architecture not only significantly improves the baseline compressed-domain models but also produces a similar performance as compared to the pixel-domain model without the need to decode the compressed-domain representations.

\section{Related Works}
\label{sec:rw}

\subsection{Compressed-Domain Computer Vision}

Some previous studies of training computer vision models were conducted in the compressed domain. Given the promising performance of learning-based compression, Torfason~\textit{et al.}~\cite{torfason2018towards} explored the use of compressed-domain representations, i.e., the quantized output of the learning-based compression encoder, to train DNNs for recognition and segmentation. Following this path, Wang \textit{et al.}~\cite{wang2022learning} jointly trained the compression models and compressed-domain computer vision models with multi-sized training images. Chen \textit{et al.}~\cite{chen2022learning} adopted the compression decoder layer to decode the compressed-domain features and trained the classification model via knowledge distillation from the pixel-domain model. Computer vision tasks such as segmentation and detection were also conducted in the compressed-domain~\cite{torfason2018towards,liu2022semantic,liu2022improving}. Other than this, some methods were proposed to operate in the traditional compressed domain, i.e., under JPEG compression~\cite{gueguen2018faster}, or in the DCT domain~\cite{xu2020learning}. These methods adopted the frequency-domain feature maps of the original uncompressed image by using the mathematically predefined DCT transform, which is not learned or optimized for compression, and their adopted frequency-domain features correspond to high-quality images with no or little compression effect. In contrast, the learning-based compressed-domain representations are deeply encoded with multiple downsampling layers and under the constraint of a rate-distortion loss during training, and in this work we adopt such compressed-domain representations at various compression bitrates.

\subsection{DNN-based Compression}

Many of the existing learning-based image coding architectures are based on DNNs~\cite{ji2023learning}. Theis \textit{et al.}~\cite{Theis2017a} defined a compressive autoencoder to learn the end-to-end trained encoder-decoder network for compression, with a joint rate-distortion loss function. Similarly, Ball\'{e} \textit{et al.}~\cite{balle2016end} described a DNN-based image compression architecture consisting of linear convolutional layers and nonlinear activation functions, and developed an end-to-end optimization framework via backpropagation for rate-distortion performance trade-off. Further improving the prior work, Ball\'{e} \textit{et al.}~\cite{balle2018variational} introduced a hyperprior network to learn and compress the spatial dependency information in the compressed-domain and adopted the MS-SSIM index into the objective function for better visual quality. Some later works were then proposed based on the hyperprior network. Lee \textit{et al.}~\cite{Lee2019Context} improved the compression performance by proposing a novel context-adaptive entropy model. Cheng \textit{et al.}~\cite{cheng2020learned} adopted ResNet blocks~\cite{he2016deep} to capture large receptive
field of the images/features in the main encoder-decoder architecture as well as the designed attention module. Gao \textit{et al.}~\cite{gao2021neural} introduced a building block to iteratively utilize the feedback residual to refine the feature maps from both low and high frequency components of the input images.

\subsection{Lightweight Attention Model}

Attention has been proposed in both natural language processing~\cite{vaswani2017attention} and computer vision~\cite{wang2018non}. The non-local block proposed in~\cite{wang2018non} is an attention model that computes the similarity of the features at different positions and selects/reinforces the positions corresponding to high similarity within each channel. Due to its relatively heavy computational cost, researchers subsequently proposed some light and efficient attention mechanisms~\cite{hu2018squeeze, woo2018cbam, cao2019gcnet, Wang_2020_CVPR}. Hu \textit{et al.}~\cite{hu2018squeeze} proposed channel-wise attention weights based on the global average pooling of output features. Woo \textit{et al.}~\cite{woo2018cbam} combined both spatial and channel attention blocks based on global average pooling as well as global maxpooling layers. Cao \textit{et al.}~\cite{cao2019gcnet} introduced a glocal context block by simplifying the non-local block of~\cite{wang2018non}. Based on~\cite{hu2018squeeze}, Wang \textit{et al.}~\cite{Wang_2020_CVPR} proposed a more parameter-efficient implementation of~\cite{hu2018squeeze} by using only one single linear layer with computations across local channels.

\section{Learning-based Compression Model}

While traditional compression algorithms usually adopt predefined and fixed linear transforms, recent DNN-based compression models learn non-linear transforms in the form of parameterized functions for encoding and decoding. This can be performed through an end-to-end rate-distortion trade-off optimization process, with the target of minimizing the cost function $R + \lambda D$, where $R$ and $D$ represent the expected compression bit rate in the compressed-domain and the distortion metric in the pixel-domain, respectively, and $\lambda$ is a weight term for $D$. In this paper, we adopt the advanced learning-based compression model with a hyperprior network~\cite{balle2018variational}, which attempts to learn the side information of the compressed-domain representations with a similar but smaller hyperprior network as compared to the compression network.

\begin{figure}[!tb]
	\centering
	\includegraphics[width=1\textwidth]{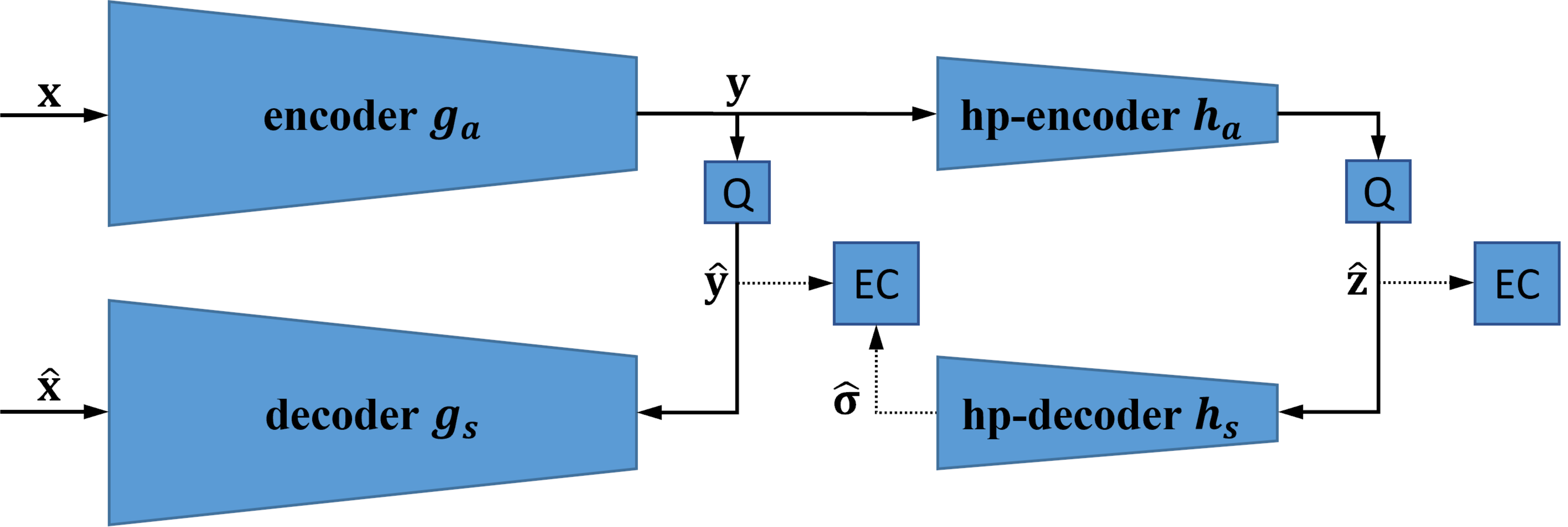}
	\caption{Block diagram of the adopted compression model. Q represents quantization and EC indicates the entropy coding process.}
	\label{fig:comp}
\end{figure}

As shown in Figure~\ref{fig:comp}, given the input image $\mathbf{x}$, the quantized compressed-domain representation $\mathbf{\hat{y}} = Q(\mathbf{y}) = Q(g_a(\mathbf{x}; \mathbf{\phi_g}))$, the quantized hyperprior compressed-domain representation $\mathbf{\hat{z}} = Q(h_a(\mathbf{y}; \mathbf{\phi_h}))$, where $g_a$/$h_a$ denotes the (hyperprior) encoder networks with the parameter sets $\mathbf{\phi_g}$/$\mathbf{\phi_h}$, respectively, and Q is the quantization function, the loss function can be expressed as follows:
\begin{equation}
\mathcal{L_C} = \log p_{\mathbf{x}|\mathbf{\hat{y}}}(\mathbf{x}|\mathbf{\hat{y}}) + \log p_{\mathbf{\hat{y}}|\mathbf{\hat{z}}}(\mathbf{\hat{y}}|\mathbf{\hat{z}}) + \log p_{\mathbf{\hat{z}}}(\mathbf{\hat{z}}).
\label{eq:comp}
\end{equation}
In Equation~\ref{eq:comp}, the first term measures a weighted distortion and the last two terms estimate the compression bit rates for $\mathbf{\hat{y}}$ and $\mathbf{\hat{z}}$, respectively. In practice, the bit rate terms can be realized by computing the representation entropy while the weighted distortion is computed using objective visual quality metrics such as MSE(PSNR) or perceptual-based metrics such as SSIM or MS-SSIM between $\mathbf{x}$ and $\mathbf{\hat{x}}$, where $\mathbf{\hat{x}} = g_s(\mathbf{\hat{y}};\mathbf{\theta_g})$ and $g_s$ is the decoder network with the parameter set $\mathbf{\theta_g}$.

\section{Proposed Compressed-Domain Visual Recognition}
\label{sec:alg}

In our work, the image classification task can be directly performed using the compressed-domain latent representation that is produced by a learning-based image codec. We adopt the variational image compression with a scale hyperprior~\cite{balle2018variational} (Figure~\ref{fig:comp}) using the MS-SSIM quality metric in the loss function (HyperMS-SSIM). Figure~\ref{fig:flow1} shows two classification paths. The first one inputs the fully decoded pixel-domain images $\mathbf{\hat{x}}$ to train/evaluate the classification model, while the other one, which is the focus of this work, directly takes the compressed-domain representation $\mathbf{\hat{y}}$ as the input of the classification model for classification, saving the computational cost by avoiding the inference time of decoding.

Given an input color image of size $D \times H' \times W'$, where $D = 3$ is the color dimension, $H'$ and $W'$ are the height and width of the image, respectively, and the compressed-domain latent representation will be of size $C \times \frac{H'}{16} \times \frac{W'}{16}$. In addition, as shown in Figure~\ref{fig:comp}, HyperMS-SSIM learns the standard deviation maps $\mathbf{\hat{\sigma}}$ to help the entropy coding process. As part of this work, we show that $\mathbf{\hat{\sigma}}$ can also improve the compressed-domain classification performance while requiring significantly less bits than the compressed-domain latent representations.

\begin{figure}[!tb]
	\centering
	\begin{subfigure}[t]{0.55\textwidth}
		\centering
		\includegraphics[width=0.9\textwidth]{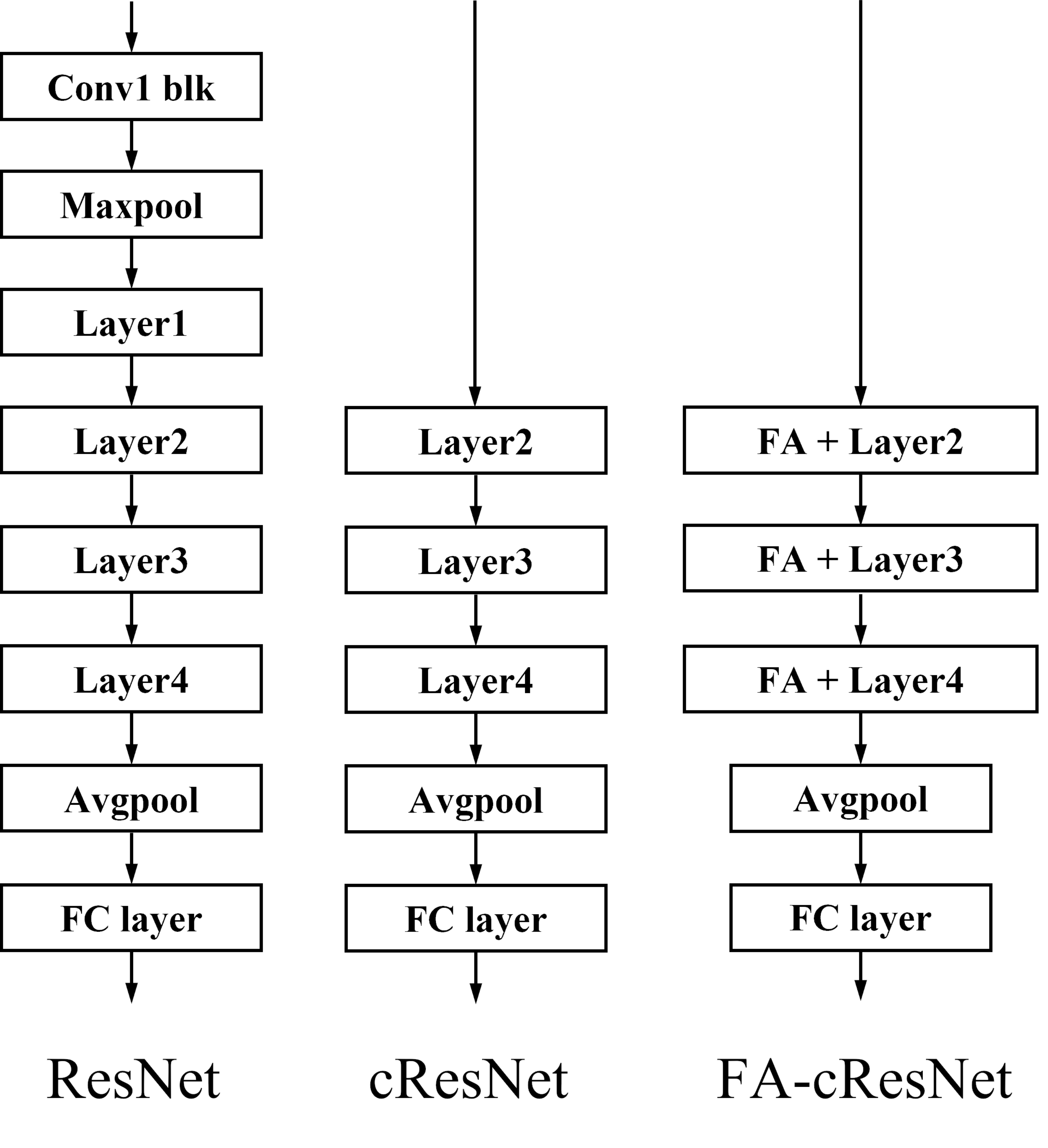}
		\caption{}
		\label{fig:arch}
	\end{subfigure}
	\begin{subfigure}[t]{0.43\textwidth}
		\centering
		\includegraphics[width=1\textwidth]{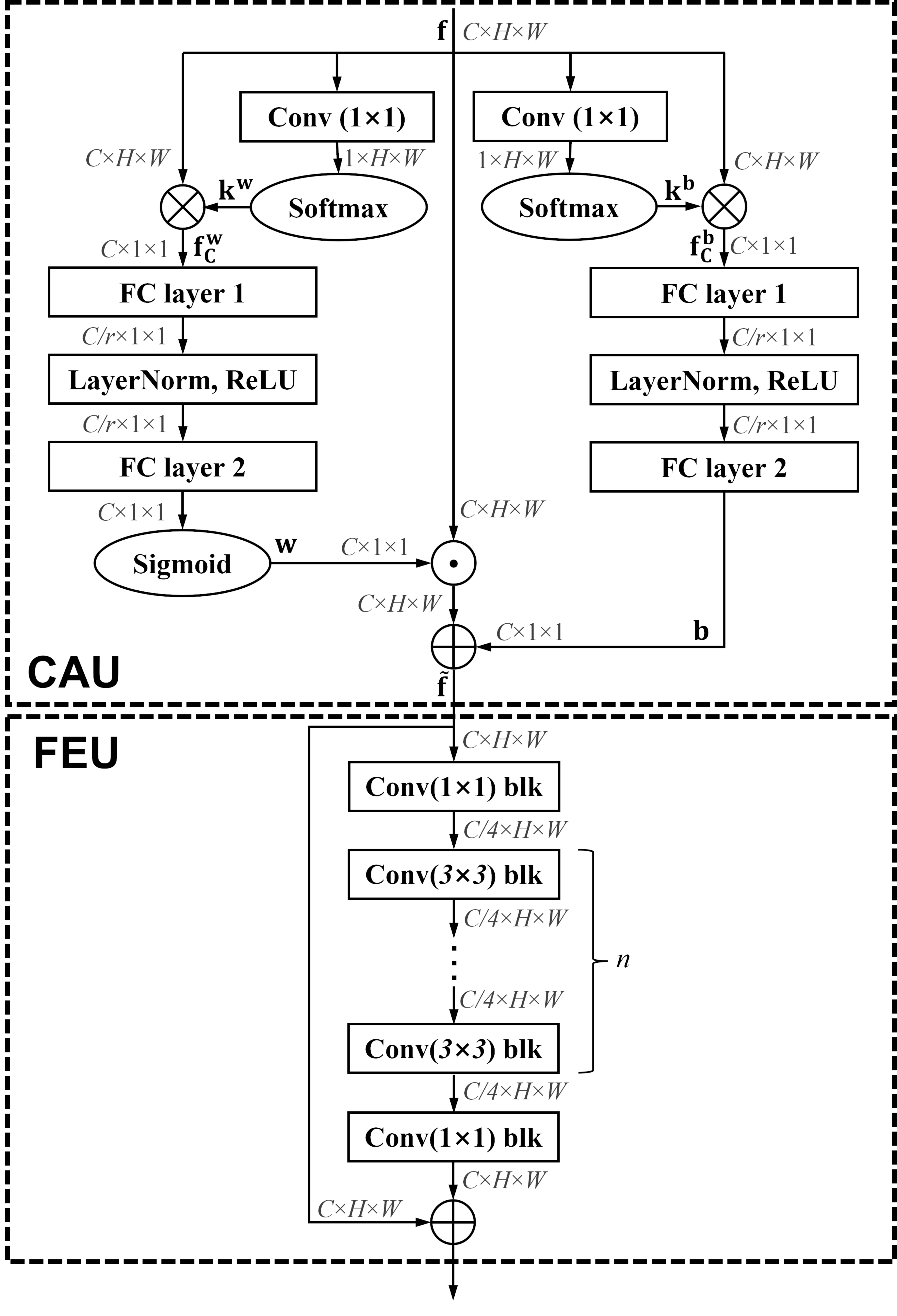}
		\caption{}
		\label{fig:fau}
	\end{subfigure}
	\caption{(a) ResNet, cResNet and our proposed FA-cResNet. (b) The proposed feature adaptation (FA) module. $\otimes$, $\odot$ and $\oplus$ indicate matrix multiplication, channel-wise multiplication and channel-wise addition, respectively. Conv ($1 \times 1$) and FC layer indicate one single $1 \times 1$ convolutional layer and fully connected layer in CAU, respectively. r is the reduction ratio in the bottleneck architecture of the two fully connected layers. In FEU, each convolutional block consists of a convolutional layer, a batch normalization layer and a ReLU function.}
\end{figure}

In \cite{torfason2018towards}, cResNet is proposed as the compressed-domain classification model by simply removing the first 11 layers of ResNet~\cite{he2016deep} to accomodate the compressed-domain representation which has a significantly reduced spatial size after encoding. Although cResNet~\cite{torfason2018towards} showed a promising direction for compressed-domain recognition, it results in a nontrivial performance drop as compared to the pixel-domain recognition on decoded images. In this section, we propose a novel feature adaptation (FA) module to enhance the compressed-domain latent representations as well as the intermediate feature maps at different cResNet stages. Figure~\ref{fig:arch} shows the architectures of ResNet, cResNet and our proposed FA-cResNet. Specifically, in our proposed FA-cResNet architecture, the first residual block of Layer 2 is replaced by our proposed FA module, which is shown in Figure~\ref{fig:fau}. Furthermore, the proposed FA module is inserted after the first residual block of Layers 3 and 4. In both cResNet and our proposed FA-cResNet, we modify the first residual block of Layer 2, i.e., the compressed-domain network head, to feed both $\mathbf{\hat{y}}$ and $\mathbf{\hat{\sigma}}$ into separate residual blocks and concatenate the outputs together, followed by one convolutional block for channel adjustment. Here $\mathbf{\hat{\sigma}}$ is adopted as an auxiliary input, whose input residual block needs a much smaller number of channels than that of $\mathbf{\hat{y}}$.

\subsection{Proposed Feature Adaptation Module}

Different from the pixel-domain images that contain redundant information both locally and globally, the compressed-domain latent representations are greedily compressed based on the target bit rate, meaning that these encoded compressed-domain features can be sparse and lack information about local contexts. A number of channels of the compressed-domain representations includes little or no non-zero values, especially for the compression models at mid to low compression rates. To avoid such insignificant channels and emphasize the relevant significant channels, an attentive feature selection procedure can be designed. As part of this work, we design an attention-based feature adaptation module incorporating a novel lightweight attention model together with a feature enhancement block. Compared to the existing lightweight attention models~\cite{hu2018squeeze,woo2018cbam,cao2019gcnet}, our feature adaptation module extends the feature weighting to a channel-wise affine transform and it further incorporates a feature enhancement block to enhance the useful features across channels. As shown in Figure~\ref{fig:fau}, the proposed FA module consists of two main units - channel-wise attention unit (CAU) and feature enhancement unit (FEU). Given a $C \times H \times W$ input feature map, where $C, H, W$ correspond to the depth, height and width dimensions of the feature map, respectively, the adaptive channel-wise affine transform is learned in the CAU based on extracted attention-based information vectors along the channels. After this, a residual-block-like FEU enhances the features across channels with cascaded learnable convolutions.

Let $\mathbf{f}$ be a $C \times H \times W$ input feature tensor. The proposed CAU is a novel affine transform along channels for selecting/reinforcing the most significant feature channels. The output of the CAU is a feature tensor as follows:
\begin{equation}
\label{eq:out}
\mathbf{\Tilde{f}} = \mathbf{w} \odot \mathbf{f} \oplus \mathbf{b},
\end{equation}
where $\odot$ and $\oplus$ denote channel-wise multiplication and channel-wise addition, respectively, and both $\mathbf{w}$ and $\mathbf{b}$ are $C \times 1 \times 1$ vectors. The channel-wise weight $\mathbf{w}$ and bias $\mathbf{b}$ are computed using bottleneck archtectures of fully connected layers~\cite{hu2018squeeze, cao2019gcnet}, i.e.,
\begin{align}
\mathbf{w} = &\ \Phi_S \circ \Phi_{FC_2}^{\mathbf{w}} \circ \Phi_{ReLU} \circ \Phi_{LN} \circ \Phi_{FC_1}^{\mathbf{w}}(\mathbf{f_C^w}),\\
\mathbf{b} = &\ \Phi_{FC_2}^{\mathbf{b}} \circ \Phi_{ReLU} \circ \Phi_{LN} \circ \Phi_{FC_1}^{\mathbf{b}}(\mathbf{f_C^b}),
\end{align}
where $\Phi_{FC_1}^{\mathbf{w}}, \Phi_{FC_2}^{\mathbf{w}}$ and $\Phi_{FC_1}^{\mathbf{b}}, \Phi_{FC_2}^{\mathbf{b}}$ represent the mapping functions of the two fully connected layers during the computation of $\mathbf{w}$ and $\mathbf{b}$, respectively, and $\Phi_{LN}, \Phi_{ReLU}$ and $\Phi_S$ indicate the mapping functions of layer normalization~\cite{ba2016layer}, ReLU activation~\cite{nair2010rectified} and sigmoid function, respectively. $\mathbf{f_C^w}$ and $\mathbf{f_C^b}$ are $C \times 1 \times 1$ initial channel-wise attention vectors that are used for the computation of $\mathbf{w}$ and $\mathbf{b}$, respectively, and which are computed using the following channel-wise attention pooling model~\cite{cao2019gcnet}:
\begin{equation}
\mathbf{f_C^m}(c) = \mathbf{k^m} \otimes \mathbf{f}(c) =\ \sum_{i=0}^{H-1}\sum_{j=0}^{W-1}\mathbf{k^m}(i,j)\mathbf{f}(c,i,j),\
\mathbf{k^m} =\ \Phi_{soft} \circ \Phi_{conv}(\mathbf{f}),\ \mathbf{m} = \mathbf{w}, \mathbf{b}
\end{equation}
where $\Phi_{conv}$ and $\Phi_{soft}$ represent the mapping functions of the $1 \times 1$ convolutional layer and softmax layer, respectively.

The transformed features $\mathbf{\Tilde{f}}$ that are output by the CAU are subsequently enhanced by the FEU. As shown in Figure~\ref{fig:fau}, the FEU consists of cascaded convolutional blocks with $1 \times 1$ convolutional blocks as the first and last blocks, and $n$ consecutive $3 \times 3$ convolutional blocks in-between. This unit is used to enhance or regenerate the features by taking into account neighboring information across channels. By increasing $n$, the output features can be produced by using the cascaded $3 \times 3$ convolutions with a larger receptive field~\cite{Simonyan15}.

The proposed feature adaptation module incorporating the CAU and FEU can improve the visual recognition performance by learning to perform an attentive enhancement of the feature maps through our designed training strategy, which will be described next.

\subsection{Training Strategy}
\label{ssec:training}
Instead of training from scratch, we adopt the weights of Layers 2-4 and the fully connected layer in the ResNet which is pretrained on pixel-domain decoded images, as the initial weights of the cResNet architecture, whose mapping function can be represented as
\begin{equation}
\Phi = \Phi_{FC} \circ \Phi_{AVG} \circ \Phi_{L_4} \circ \Phi_{L_3} \circ \Phi_{L_2},
\end{equation}
where $\Phi_{AVG}$ and $\Phi_{FC}$ represent the mapping functions of average pooling and the fully connected layer, respectively, and
\begin{equation}
\Phi_{L_i} = \Phi_{N_i-1, L_i} \circ ... \circ \Phi_{1, L_i} \circ \Phi_{0, L_i}, \quad i = 2, 3, 4,
\end{equation}
where $\Phi_{L_i}$ indicates the overall mapping function of Layer $i$ ($i = 2, 3, 4$), which includes the mapping functions $\Phi_{j, L_i}$, ($j = 0, 1, ... , N_i - 1$) for $N_i$ residual blocks.

The FA module mapping function can be represented as $\Phi_{FA, L_i}$, where $i$ indicates the index of Layers 2, 3 and 4. Note that since we use both the compressed-domain representations $\mathbf{\hat{y}}$ and the standard deviation maps $\mathbf{\hat{\sigma}}$ as input, we replace the first residual block of Layer 2 in FA-cResNet with two parallel FA modules $\Phi_{FA_\mathbf{\hat{y}}}$ and $\Phi_{FA_\mathbf{\hat{\sigma}}}$ for $\mathbf{\hat{y}}$ and $\mathbf{\hat{\sigma}}$, respectively. Furthermore, we insert our proposed FA module in Layers 3 and 4 to build our FA-cResNet. Then we freeze all the adopted pretrained weights and only update the inserted FA modules during training. The mapping function now becomes\footnote{Average pooling is a nonparametric mapping function so it does not need updating/adapting.}
\begin{align}
&\Phi^{FA} = \Phi_{FC} \circ \Phi_{AVG} \circ \Phi_{L_4}^{FA} \circ \Phi_{L_3}^{FA} \circ \Phi_{L_2}^{FA},\\
&\Phi_{L_2}^{FA} = \Phi_{N_2-1, L_2} \circ ... \circ \Phi_{1, L_2} \circ (\Phi_{FA_\mathbf{\hat{y}}, L_2}^{T} + \Phi_{FA_\mathbf{\hat{\sigma}}, L_2}^{T}),\\
&\Phi_{L_i}^{FA} = \Phi_{N_i-1, L_i} \circ ... \circ \Phi_{1, L_i} \circ \Phi_{FA, L_i}^{T} \circ \Phi_{0, L_i}, \quad i = 3, 4,
\end{align}
where the $T$ superscript is used to denote that the FA modules are the only ones that are being trained, and $+$ denotes a concatenation along channels of the input block mappings $\Phi_{FA_\mathbf{\hat{y}}, L_2}^{T}$ and $\Phi_{FA_\mathbf{\hat{\sigma}}, L_2}^{T}$. After the partial training process reaches its plateau, all weights in our FA-cResNet are then unfreezed and finetuned as follows:
\begin{align}
&\Phi^{FT} = \Phi_{FC}^{FT} \circ \Phi_{AVG} \circ \Phi_{L_4}^{FT} \circ \Phi_{L_3}^{FT} \circ \Phi_{L_2}^{FT},\\
&\Phi_{L_2}^{FT} = \Phi_{N_2-1, L_2}^{FT} \circ ... \circ \Phi_{1, L_2}^{FT} \circ (\Phi_{FA_\mathbf{\hat{y}}, L_2}^{FT} + \Phi_{FA_\mathbf{\hat{\sigma}}, L_2}^{FT}),\\
&\Phi_{L_i}^{FT} = \Phi_{N_i-1, L_i}^{FT} \circ ... \circ \Phi_{1, L_i}^{FT} \circ \Phi_{FA, L_i}^{FT} \circ \Phi_{0, L_i}^{FT}, \quad i = 3, 4,
\end{align}
where $FT$ indicates that all the trainable mapping functions have been finetuned using the compressed-domain latent representations $\mathbf{\hat{y}}$ and the standard deviation maps $\mathbf{\hat{\sigma}}$ corresponding to the images in the training dataset.

Given the trainable parameters $\mathbf{\theta_r}$ in the FA-cResNet, our loss function can be expressed as:
\begin{equation}
\mathcal{L}(\mathbf{\theta_r}) = \frac{1}{N}\sum_{i=0}^{N - 1}\mathcal{L}_{ce}(t_i, \Phi^{FA}(\mathbf{\hat{y}}_i,\mathbf{\hat{\sigma}}_i;\theta_r)) + \lambda \mathcal{R}(\mathbf{\theta_r}),
\end{equation}
where $\mathcal{L}_{ce}$ represents the cross-entropy loss, $t_i$ denotes the target groundtruth label for the $i^{th}$ input features ($\mathbf{\hat{y}}_i,\mathbf{\hat{\sigma}}_i$) and $\mathcal{R}$ is a regularizer term with a coefficient $\lambda$.

\section{Experimental Results}
\label{sec:exp}

\begin{figure}[!tb]
	\centering
	\begin{minipage}{0.16\textwidth}
		\centering
		brick
	\end{minipage}
	\begin{minipage}{0.16\textwidth}
		\centering
		foliage
	\end{minipage}
	\begin{minipage}{0.16\textwidth}
		\centering
		leather
	\end{minipage}
	\begin{minipage}{0.16\textwidth}
		\centering
		wallpaper
	\end{minipage}
	\begin{minipage}{0.16\textwidth}
		\centering
		water
	\end{minipage}
	\begin{minipage}{0.16\textwidth}
		\centering
		wood
	\end{minipage}\\
	\scriptsize
	\begin{minipage}{0.163\textwidth}
		\centering
		\includegraphics[width=1\textwidth]{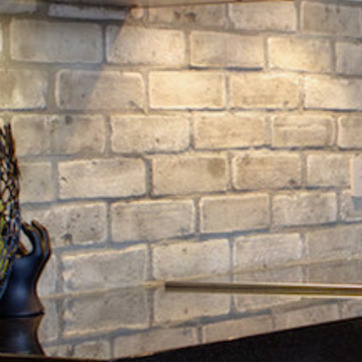}\\
		bit rate: 2.551 bpp\\
	\end{minipage}
	\begin{minipage}{0.163\textwidth}
		\centering
		\includegraphics[width=1\textwidth]{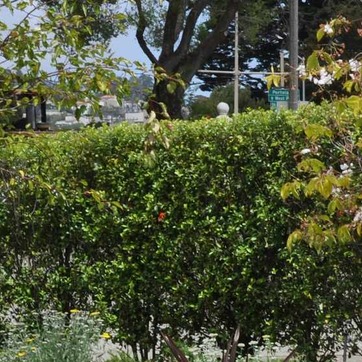}\\
		bit rate: 4.872 bpp\\
	\end{minipage}
	\begin{minipage}{0.163\textwidth}
		\centering
		\includegraphics[width=1\textwidth]{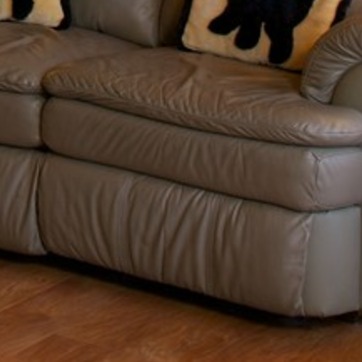}\\
		bit rate: 1.444 bpp\\
	\end{minipage}
	\begin{minipage}{0.163\textwidth}
		\centering
		\includegraphics[width=1\textwidth]{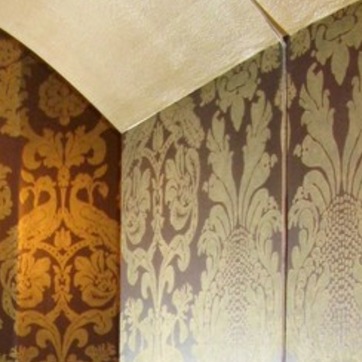}\\
		bit rate: 2.228 bpp\\
	\end{minipage}
	\begin{minipage}{0.163\textwidth}
		\centering
		\includegraphics[width=1\textwidth]{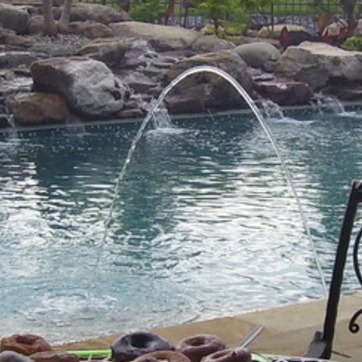}\\
		bit rate: 3.132 bpp\\
	\end{minipage}
	\begin{minipage}{0.163\textwidth}
		\centering
		\includegraphics[width=1\textwidth]{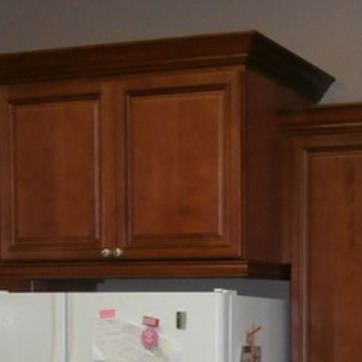}\\
		bit rate: 1.223 bpp\\
	\end{minipage}\\
	\ \\
	\ \\
	\begin{minipage}{0.163\textwidth}
		\centering
		\includegraphics[width=1\textwidth]{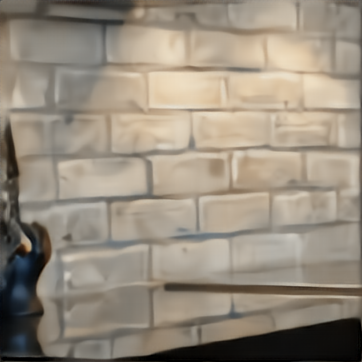}
		bit rate: 0.108 bpp\\
		MS-SSIM: 0.9282\\
		\includegraphics[width=0.315\textwidth]{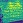}
		\includegraphics[width=0.315\textwidth]{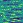}
		\includegraphics[width=0.315\textwidth]{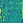}
		\includegraphics[width=0.315\textwidth]{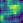}
		\includegraphics[width=0.315\textwidth]{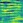}
		\includegraphics[width=0.315\textwidth]{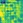}
	\end{minipage}
	\begin{minipage}{0.163\textwidth}
		\centering
		\includegraphics[width=1\textwidth]{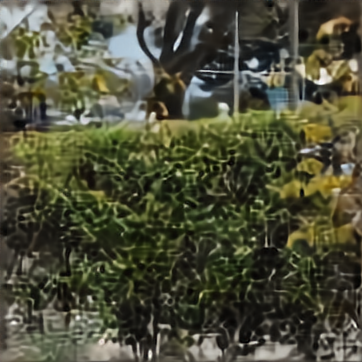}
		bit rate: 0.187 bpp\\
		MS-SSIM: 0.8560\\
		\includegraphics[width=0.315\textwidth]{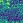}
		\includegraphics[width=0.315\textwidth]{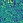}
		\includegraphics[width=0.315\textwidth]{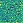}
		\includegraphics[width=0.315\textwidth]{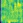}
		\includegraphics[width=0.315\textwidth]{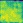}
		\includegraphics[width=0.315\textwidth]{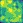}
	\end{minipage}
	\begin{minipage}{0.163\textwidth}
		\centering
		\includegraphics[width=1\textwidth]{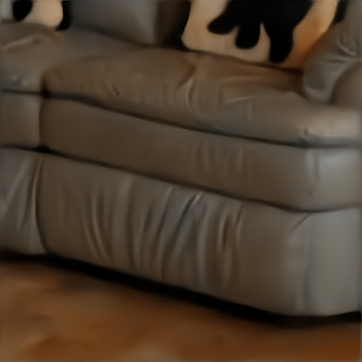}
		bit rate: 0.0819 bpp\\
		MS-SSIM: 0.9491\\
		\includegraphics[width=0.315\textwidth]{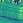}
		\includegraphics[width=0.315\textwidth]{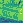}
		\includegraphics[width=0.315\textwidth]{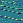}
		\includegraphics[width=0.315\textwidth]{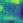}
		\includegraphics[width=0.315\textwidth]{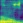}
		\includegraphics[width=0.315\textwidth]{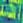}
	\end{minipage}
	\begin{minipage}{0.163\textwidth}
		\centering
		\includegraphics[width=1\textwidth]{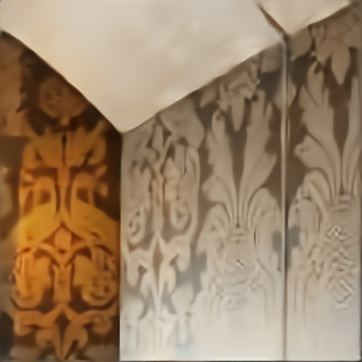}
		bit rate: 0.118 bpp\\
		MS-SSIM: 0.9206\\
		\includegraphics[width=0.315\textwidth]{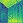}
		\includegraphics[width=0.315\textwidth]{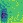}
		\includegraphics[width=0.315\textwidth]{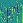}
		\includegraphics[width=0.315\textwidth]{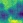}
		\includegraphics[width=0.315\textwidth]{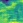}
		\includegraphics[width=0.315\textwidth]{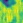}
	\end{minipage}
	\begin{minipage}{0.163\textwidth}
		\centering
		\includegraphics[width=1\textwidth]{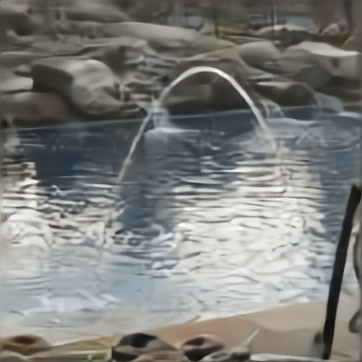}
		bit rate: 0.141 bpp\\
		MS-SSIM: 0.9097\\
		\includegraphics[width=0.315\textwidth]{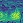}
		\includegraphics[width=0.315\textwidth]{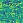}
		\includegraphics[width=0.315\textwidth]{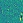}
		\includegraphics[width=0.315\textwidth]{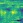}
		\includegraphics[width=0.315\textwidth]{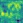}
		\includegraphics[width=0.315\textwidth]{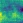}
	\end{minipage}
	\begin{minipage}{0.163\textwidth}
		\centering
		\includegraphics[width=1\textwidth]{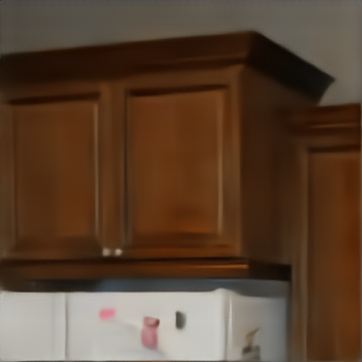}
		bit rate: 0.0738 bpp\\
		MS-SSIM: 0.9645\\
		\includegraphics[width=0.315\textwidth]{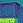}
		\includegraphics[width=0.315\textwidth]{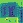}
		\includegraphics[width=0.315\textwidth]{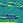}
		\includegraphics[width=0.315\textwidth]{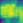}
		\includegraphics[width=0.315\textwidth]{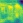}
		\includegraphics[width=0.315\textwidth]{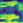}
	\end{minipage}\\
	\ \\
	\ \\
	\begin{minipage}{0.163\textwidth}
		\centering
		\includegraphics[width=1\textwidth]{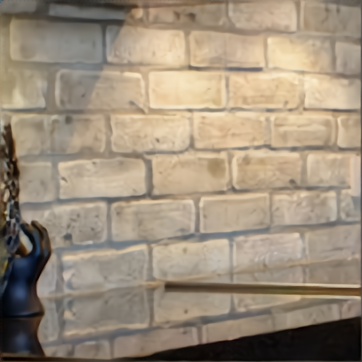}
		bit rate: 0.385 bpp\\
		MS-SSIM: 0.9854\\
		\includegraphics[width=0.315\textwidth]{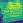}
		\includegraphics[width=0.315\textwidth]{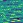}
		\includegraphics[width=0.315\textwidth]{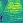}
		\includegraphics[width=0.315\textwidth]{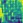}
		\includegraphics[width=0.315\textwidth]{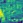}
		\includegraphics[width=0.315\textwidth]{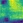}
	\end{minipage}
	\begin{minipage}{0.163\textwidth}
		\centering
		\includegraphics[width=1\textwidth]{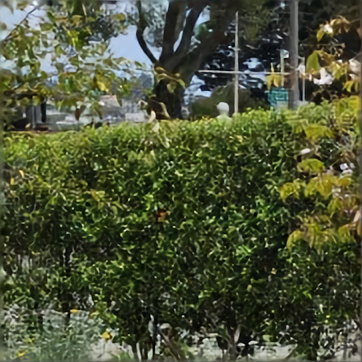}
		bit rate: 0.737 bpp\\
		MS-SSIM: 0.9740\\
		\includegraphics[width=0.315\textwidth]{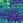}
		\includegraphics[width=0.315\textwidth]{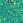}
		\includegraphics[width=0.315\textwidth]{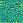}
		\includegraphics[width=0.315\textwidth]{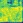}
		\includegraphics[width=0.315\textwidth]{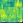}
		\includegraphics[width=0.315\textwidth]{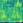}
	\end{minipage}
	\begin{minipage}{0.163\textwidth}
		\centering
		\includegraphics[width=1\textwidth]{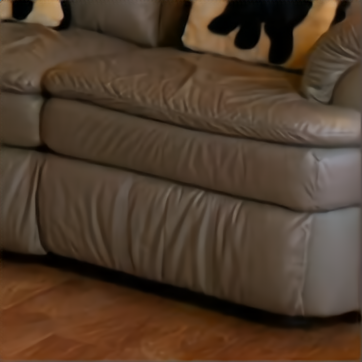}
		bit rate: 0.265 bpp\\
		MS-SSIM: 0.9890\\
		\includegraphics[width=0.315\textwidth]{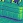}
		\includegraphics[width=0.315\textwidth]{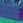}
		\includegraphics[width=0.315\textwidth]{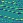}
		\includegraphics[width=0.315\textwidth]{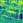}
		\includegraphics[width=0.315\textwidth]{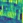}
		\includegraphics[width=0.315\textwidth]{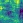}
	\end{minipage}
	\begin{minipage}{0.163\textwidth}
		\centering
		\includegraphics[width=1\textwidth]{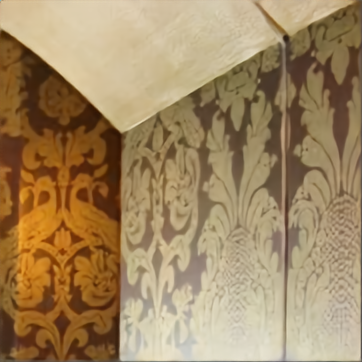}
		bit rate: 0.431 bpp\\
		MS-SSIM: 0.9840\\
		\includegraphics[width=0.315\textwidth]{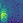}
		\includegraphics[width=0.315\textwidth]{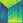}
		\includegraphics[width=0.315\textwidth]{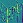}
		\includegraphics[width=0.315\textwidth]{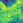}
		\includegraphics[width=0.315\textwidth]{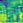}
		\includegraphics[width=0.315\textwidth]{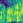}
	\end{minipage}
	\begin{minipage}{0.163\textwidth}
		\centering
		\includegraphics[width=1\textwidth]{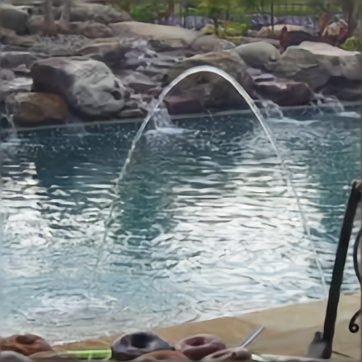}
		bit rate: 0.531 bpp\\
		MS-SSIM: 0.9854\\
		\includegraphics[width=0.315\textwidth]{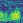}
		\includegraphics[width=0.315\textwidth]{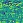}
		\includegraphics[width=0.315\textwidth]{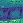}
		\includegraphics[width=0.315\textwidth]{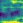}
		\includegraphics[width=0.315\textwidth]{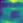}
		\includegraphics[width=0.315\textwidth]{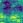}
	\end{minipage}
	\begin{minipage}{0.163\textwidth}
		\centering
		\includegraphics[width=1\textwidth]{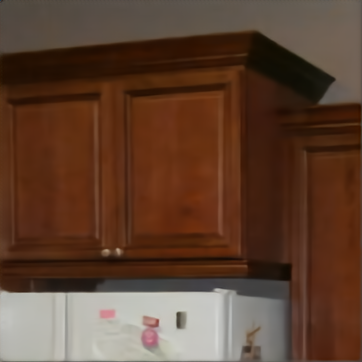}
		bit rate: 0.202 bpp\\
		MS-SSIM: 0.9892\\
		\includegraphics[width=0.315\textwidth]{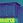}
		\includegraphics[width=0.315\textwidth]{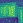}
		\includegraphics[width=0.315\textwidth]{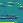}
		\includegraphics[width=0.315\textwidth]{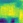}
		\includegraphics[width=0.315\textwidth]{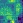}
		\includegraphics[width=0.315\textwidth]{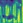}
	\end{minipage}\\
	\ \\
	\ \\
	\begin{minipage}{0.163\textwidth}
		\centering
		\includegraphics[width=1\textwidth]{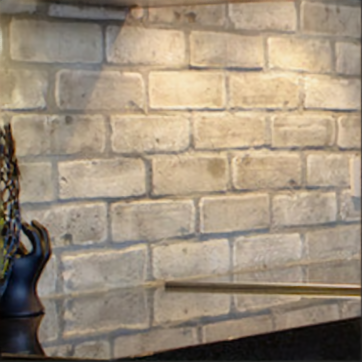}
		bit rate: 1.415 bpp\\
		MS-SSIM: 0.9980\\
		\includegraphics[width=0.315\textwidth]{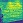}
		\includegraphics[width=0.315\textwidth]{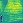}
		\includegraphics[width=0.315\textwidth]{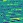}
		\includegraphics[width=0.315\textwidth]{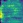}
		\includegraphics[width=0.315\textwidth]{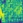}
		\includegraphics[width=0.315\textwidth]{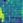}
	\end{minipage}
	\begin{minipage}{0.163\textwidth}
		\centering
		\includegraphics[width=1\textwidth]{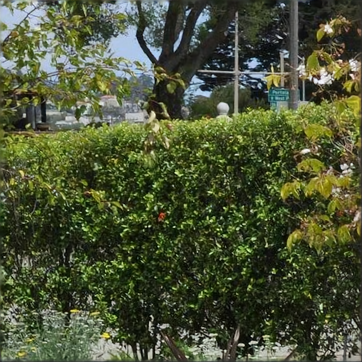}
		bit rate: 2.974 bpp\\
		MS-SSIM: 0.9973\\
		\includegraphics[width=0.315\textwidth]{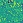}
		\includegraphics[width=0.315\textwidth]{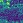}
		\includegraphics[width=0.315\textwidth]{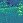}
		\includegraphics[width=0.315\textwidth]{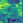}
		\includegraphics[width=0.315\textwidth]{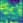}
		\includegraphics[width=0.315\textwidth]{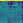}
	\end{minipage}
	\begin{minipage}{0.163\textwidth}
		\centering
		\includegraphics[width=1\textwidth]{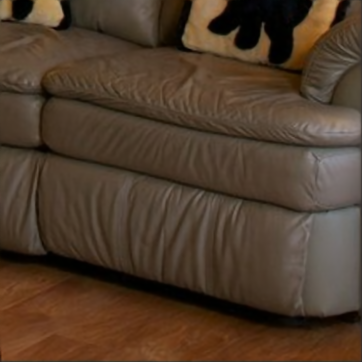}
		bit rate: 0.953 bpp\\
		MS-SSIM: 0.9977\\
		\includegraphics[width=0.315\textwidth]{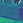}
		\includegraphics[width=0.315\textwidth]{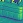}
		\includegraphics[width=0.315\textwidth]{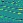}
		\includegraphics[width=0.315\textwidth]{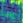}
		\includegraphics[width=0.315\textwidth]{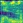}
		\includegraphics[width=0.315\textwidth]{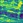}
	\end{minipage}
	\begin{minipage}{0.163\textwidth}
		\centering
		\includegraphics[width=1\textwidth]{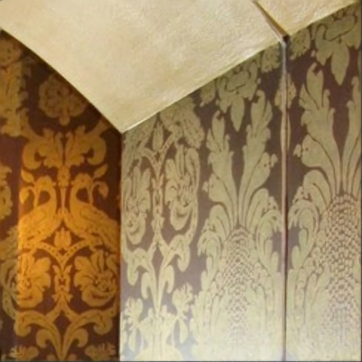}
		bit rate: 1.548 bpp\\
		MS-SSIM: 0.9976\\
		\includegraphics[width=0.315\textwidth]{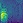}
		\includegraphics[width=0.315\textwidth]{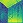}
		\includegraphics[width=0.315\textwidth]{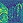}
		\includegraphics[width=0.315\textwidth]{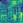}
		\includegraphics[width=0.315\textwidth]{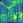}
		\includegraphics[width=0.315\textwidth]{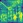}
	\end{minipage}
	\begin{minipage}{0.163\textwidth}
		\centering
		\includegraphics[width=1\textwidth]{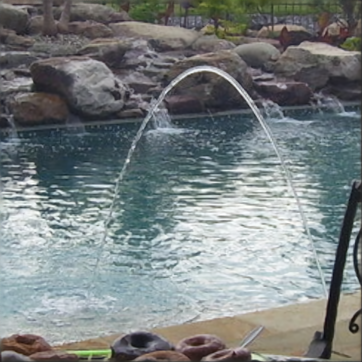}
		bit rate: 1.866 bpp\\
		MS-SSIM: 0.9983\\
		\includegraphics[width=0.315\textwidth]{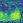}
		\includegraphics[width=0.315\textwidth]{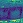}
		\includegraphics[width=0.315\textwidth]{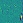}
		\includegraphics[width=0.315\textwidth]{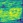}
		\includegraphics[width=0.315\textwidth]{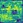}
		\includegraphics[width=0.315\textwidth]{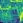}
	\end{minipage}
	\begin{minipage}{0.163\textwidth}
		\centering
		\includegraphics[width=1\textwidth]{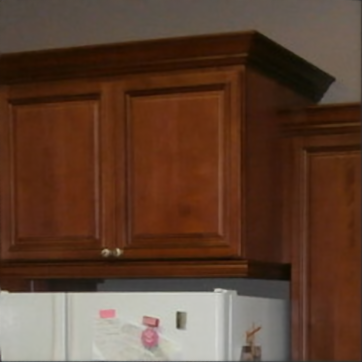}
		bit rate: 0.872 bpp\\
		MS-SSIM: 0.9973\\
		\includegraphics[width=0.315\textwidth]{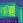}
		\includegraphics[width=0.315\textwidth]{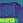}
		\includegraphics[width=0.315\textwidth]{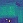}
		\includegraphics[width=0.315\textwidth]{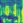}
		\includegraphics[width=0.315\textwidth]{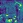}
		\includegraphics[width=0.315\textwidth]{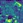}
	\end{minipage}\\
	\caption{Examples of texture images from the MINC-2500 dataset~\cite{bell15minc} (row 1) and corresponding decoded images under the compression models HyperMS-SSIM-1 (row 2), HyperMS-SSIM-4 (row 3) and HyperMS-SSIM-8 (row 4). The bit rate is given below each image. Under each decoded image, we also show the MS-SSIM values as well as three most informative channels of $\mathbf{\hat{y}}$ and $\mathbf{\hat{\sigma}}$, which are measured by the number of non-zero values within each channel.}
	\label{fig:eg}
\end{figure}

\subsection{Datasets and Models}

In this section, we implement the proposed FA-cResNet on the ImageNet dataset~\cite{deng2009imagenet} for natural image recognition and the Materials in Context (MINC) Database~\cite{bell15minc} for texture recognition. ImageNet is a well-known benchmark with 1000 classes for natural image recognition, including more than 1.2 million training images and 50,000 images in the validation set. MINC-2500 is a real-world texture recognition dataset, and each image contains not only the key object but also the context/background in which the object lies. Specifically, its subset MINC-2500 is adopted. MINC-2500 contains 57,500 images of 23 classes. Each class contains 2,500 images. We use the train-validation-test split 1 provided in the MINC-2500 dataset.

We employ the HyperMS-SSIM tensorflow-compression implementation available at~\cite{tfc}, which provides 8 trained models for HyperMS-SSIM corresponding to different quality levels, and thus rates, named as “bmshj2018-hyperprior-MS-SSIM-[1-8]” (1 for lowest quality/rate and 8 for highest quality/rate). We pick the models 1, 3, 4, 5 under similar compression rates with existing methods for comparison on ImageNet, and choose models 1, 4, 8 which correspond to the lowest, middle, and highest quality/rate, respectively, on all images in the MINC-2500 dataset. Some examples of (decoded) images as well as compressed-domain features are given in Figure~\ref{fig:eg}. We build our FA-cResNet model by modifying the cResNet-39 model, which is obtained by removing the first 11 layers of ResNet-50 as in~\cite{torfason2018towards}, and compare their performances.

\subsection{Anchor Training}

For performance comparison, we define two different anchors which are performed on either the decoded images or the original images. The decoded images are obtained by using the HyperMS-SSIM models to encode the original images and then fully decode the obtained compressed-domain representations. The considered anchors are described below:
\begin{itemize}
	\item \textbf{Anchor 1} - The classification task is performed directly by applying a ResNet-50 model that is pretrained on the original uncompressed image dataset.
	\item \textbf{Anchor 2} – The classification task is performed by applying a ResNet-50 model that is retrained using decoded images from the considered image dataset.
\end{itemize}

The input images are resized to $256 \times 256$ with bilinear interpolation and then cropped to $224 \times 224$ with the Pytorch random resized crop function, followed by a random horizontal flipping. For ImageNet~\cite{deng2009imagenet}, the anchor ResNet-50~\cite{he2016deep} models (Anchor 2) are finetuned based on the Pytorch ResNet-50 model (Anchor 1) over the decoded ImageNet training set with a batch size of 64. We use a stochastic gradient descent (SGD) optimizer with a learning rate of 0.0001, a momentum of 0.9 and a weight decay of 0.0001 for 5 epochs. For MINC-2500~\cite{bell15minc}, all anchor ResNet-50 classifiers are trained from scratch on either the original (Anchor 1) or the decoded (Anchor 2) MINC-2500 training set. The batch size is set to 32. An SGD optimizer with a momentum of 0.9 and a weight decay of 0.001 is used. The learning rate is initialized to 0.01 and divided by a factor of 10 after 60 and 80 epochs for a total of 90 epochs.

\subsection{Compressed-Domain Training}

The input compressed-domain representations $\mathbf{\hat{y}}$ and standard deviation maps $\mathbf{\hat{\sigma}}$ are both randomly cropped and then resized to $28 \times 28$ with nearest-neighbor interpolation by the Pytorch random resized crop function which is suggested in~\cite{zhang2017deep}. We adopt the weights of the pretrained pixel-domain anchor model and follow the training strategy as described in Section~\ref{ssec:training}. For training the natural image recognition models, the pretrained layers are first frozen for 15 epochs and then finetuned with a learning rate of 0.0001. The initial learning rate of the inserted FA modules is set to 0.01 with a learning rate decay of 0.1 after 10 and 25 epochs. The training stops after 35 epochs. We use a batch size of 64 and an SGD optimizer with a momentum of 0.9 and a weight decay of 0.0005. For training the texture recognition models, the adopted layers frozen in the first 20 epochs, and then the whole network is optimized in the remaining epochs. The learning rate is initialized to 0.01 and divided by a factor of 10 after 40 and 60 epochs for a total of 70 epochs. A batch size of 32 and an SGD optimizer with a momentum of 0.9 and a weight decay of 0.001 are used.

We use a stride of 1 in the input blocks of both cResNet~\cite{torfason2018towards} and FA-cResNet to preserve the spatial size. The number of output channels of the input blocks are set to 512 for $\mathbf{\hat{y}}$ and 4 for $\mathbf{\hat{\sigma}}$. The output feature maps are concatenated and mapped to a 512-channel feature map using a convolutional block. For the parameters in the FA module, the reduction ratio $r$ in CAU is set to 16, and the number of $3 \times 3$ convolutional blocks $n$ in the FEU is chosen to be 2/3/4 for HyperMS-SSIM compression model 1 and 3/4/5 for other compression models in the inserted FA modules of Layer 2/3/4, respectively.

\subsection{Performance Comparison on ImageNet}

\begin{figure*}[!t]
	\centering
	\begin{minipage}{0.32\textwidth}
		\centering
		\includegraphics[width=1\textwidth]{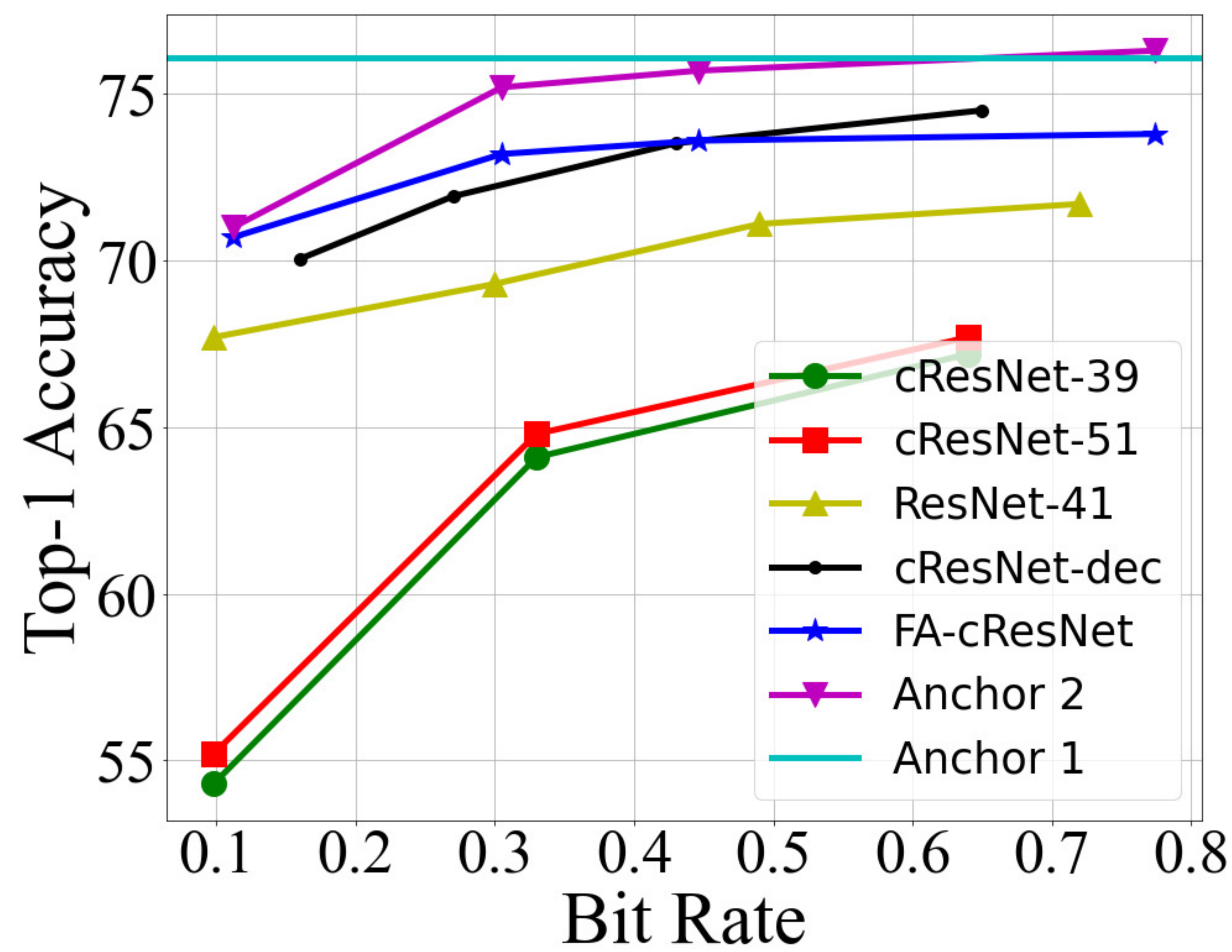}
	\end{minipage}
	\begin{minipage}{0.32\textwidth}
		\centering
		\includegraphics[width=1\textwidth]{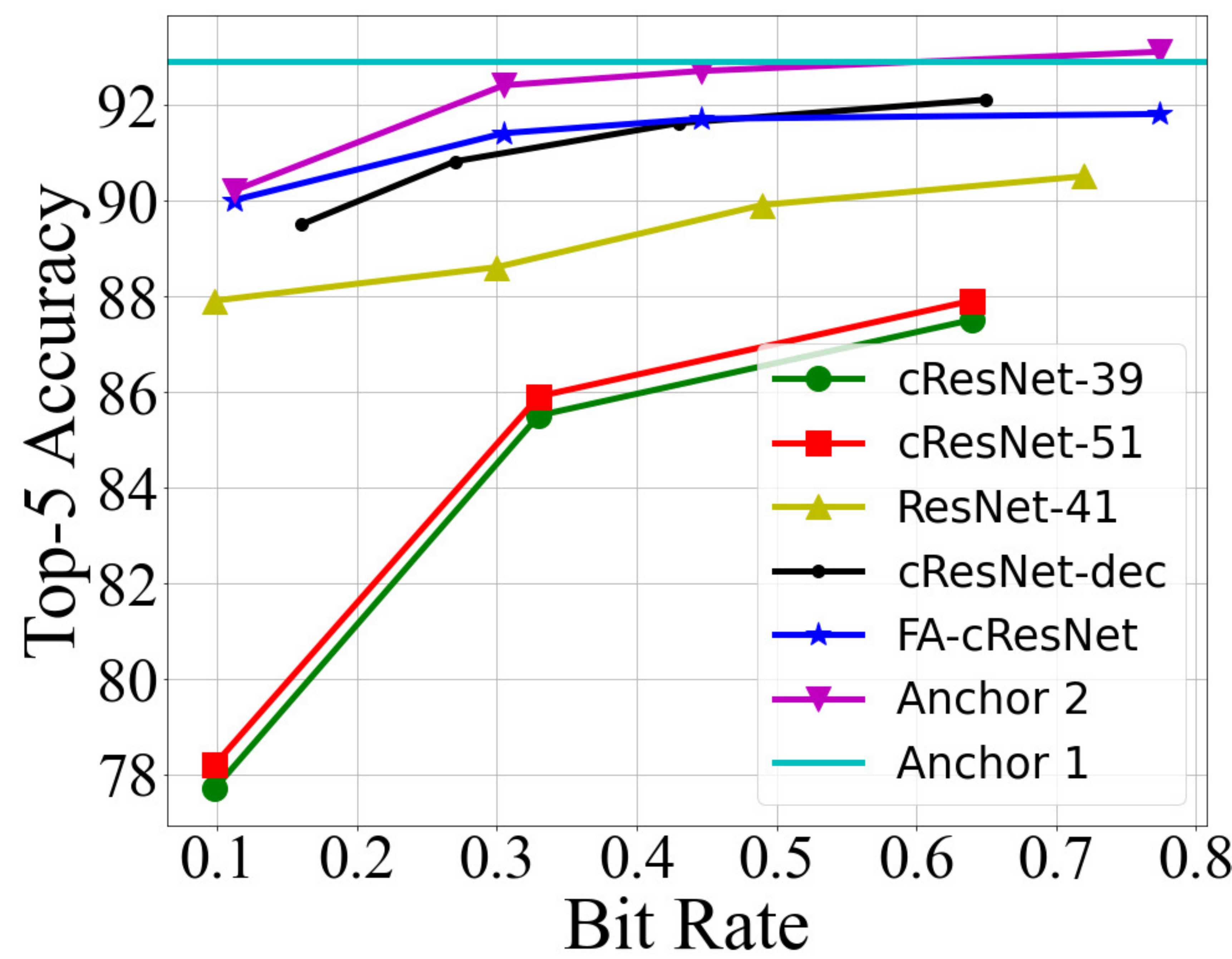}
	\end{minipage}
	\caption{Curves of accuracy results vs. bit rate on the ImageNet dataset. The bit rates are averaged over the ImageNet validation dataset.}
	\label{fig:imgnet}
\end{figure*}

We first compare our method with other existing work as well as the anchor models on the ImageNet dataset. To the best of our knowledge, \cite{torfason2018towards} and \cite{wang2022learning} are the two existing methods that directly adopt compressed-domain representations that are generated by the learning-based compression encoder as the input to train the compressed-domain classification model without decoding. We also show the results from~\cite{chen2022learning} which takes the partial decoded features of the compressed-domain representations as input. We adopt the compressed-domain features under bit rates that are similar to the ones adopted in the existing work for comparison. 

Since we only use the compressed-domain representations of single-sized original images in the ImageNet dataset, we compare our results with the published results in~\cite{torfason2018towards} (cResNet-39 and cResNet-51) and~\cite{wang2022learning} (ResNet-41) on single-sized image training. As shown in Figure~\ref{fig:imgnet}, our proposed FA-cResNet can outperform the methods of~\cite{torfason2018towards} and~\cite{wang2022learning} in terms of higher Top-1 and Top-5 accuracy results. Compared to ResNet-41, our method yields higher computational efficiency given that we discard the compression encoder and use only the extracted compressed-domain features during training while the encoding process is required for training ResNet-41 with the preprocessed original images as the input. Compared to~\cite{chen2022learning} (cResNet-dec), our method produces higher accuracy results under relatively low compression rates ($\le$ 0.4 bpp) while performing a little worse at high compression rates. However, \cite{chen2022learning} takes advantage of decoded layers in the compression model to deconvolve the compressed-domain representations, and our training and inference are performed offline from the compression and without the need of any decoding. Although there exists a small accuracy performance gap between our FA-cResNet and Anchor 2, our FA-cResNet needs significantly less computational time by avoiding the decoding process, which will be illustrated in Section~\ref{ssec:eff}.

\subsection{Classification Performance on the MINC-2500 Texture Dataset}

\begin{table}[!tb]
	\centering
	\caption{Top-1 accuracy results on the MINC-2500 dataset using the proposed FA-cResNet at various qualities/rates and comparisons with cResNet and pixel-domain anchors. Bold numbers indicate best performance.}
	\label{tab:acc}
	\renewcommand{\arraystretch}{1.2}
	\resizebox{0.8\textwidth}{!}{
		\begin{tabular}{c||ccc|c}
			\bottomrule[2pt]
			\textbf{Anchor 1}&\multirow{2}{*}{\textbf{Anchor 2}}&\multirow{2}{*}{\textbf{cResNet}}&\multirow{2}{*}{\textbf{FA-cResNet}}&\\
			(2.604 bpp)&&&&\\
			\hline
			\multirow{3}{*}{75.7}&71.8&70.3&\textbf{73.2}&\textbf{HyperMS-SSIM-1} (0.0888 bpp)\\
			&75.3&72.9&\textbf{76.5}&\textbf{HyperMS-SSIM-4} (0.309 bpp)\\
			&75.9&74.4&\textbf{77.9}&\textbf{HyperMS-SSIM-8} (1.366 bpp)\\
			\toprule[2pt]
	\end{tabular}}
\end{table}

Although many variants based on some classic backbone model such as ResNet~\cite{he2016deep} have been proposed for texture recognition, a recent study~\cite{deng2020study} shows that the vanilla ResNet model yields good performance on the texture recognition task and the modifications~\cite{zhang2017deep, xue2018deep, hu2019multi} based on the ResNet backbone result in trivial improvements or even worse performances~\cite{deng2020study}. So in this paper we only consider ResNet as our baseline model. We train the cResNet model as baseline for comparison with our FA-cResNet. Similar to FA-cResNet, we replace the first residual block of cResNet with two parallel residual blocks to accommodate the inputs $\mathbf{\hat{y}}$ and $\mathbf{\hat{\sigma}}$. Table~\ref{tab:acc} shows the Top-1 accuracy results for Anchors 1 and 2, cResNet and our FA-cResNet based on the MINC-2500 test dataset with the original images (Anchor 1), decoded images (Anchor 2) and compressed-domain representations (cResNet/FA-cResNet), the last two of which are produced by each of HyperMS-SSIM-1, HyperMS-SSIM-4 and HyperMS-SSIM-8. We also list the average bit rate (bbp) over all test images for each compression model.

From Table~\ref{tab:acc}, it can be seen that, at the lowest and mid quality level/bit rate (HyperMS-SSIM-1 and HyperMS-SSIM-4), there is a clear decrease in classification performance for Anchor 2. The compressed-domain classification using cResNet yields a lower classification performance in terms of Top-1 accuracy as compared to both anchors. For each of the adopted compression rates, our FA-cResNet clearly improves the performance on compressed-domain classification by approximately 3\% in terms of Top-1 accuracy as compared to the baseline cResNet model, and also produces a better performance than each corresponding Anchor 2 model. Moreover, at the mid and high quality levels/bit rates, our FA-cResNet results in the highest Top-1 accuracy result as compared to cResNet and to all the anchor models.

\subsection{Computational Efficiency}
\label{ssec:eff}
\begin{table}[!tb]
	\centering
	\caption{The average single-input inference time (inf. time), the total feature size during forward/backward pass (pass size), the number of parameters (\#params) and floating point operations (FLOPs) for Anchor 2 and the proposed FA-cResNet, and the average single-input encoding time and decoding time under each of the three compression models. Results are produced using the MINC-2500 dataset. Bold number indicates better performance.}
	\label{tab:eff}
	\renewcommand{\arraystretch}{1.2}
	\resizebox{1\textwidth}{!}{
		\begin{tabular}{c|cccccc|c}
			\bottomrule[2pt]
			&\textbf{enc. time}&\textbf{dec. time}&\textbf{inf. time $\downarrow$}&\textbf{pass size $\downarrow$}&\textbf{\#params $\downarrow$}&\textbf{FLOPs $\downarrow$}&\\
			\hline
			\multirow{2}{*}{\textbf{HyperMS-SSIM-1}}&\multirow{2}{*}{478 ms}&296 ms&2.2 ms&177.8 M&\textbf{23.6 M}&\textbf{4.1 G}&\textbf{Anchor 2}\\
			&&-&\textbf{1.8 ms}&\textbf{107.8 M}&38.9 M&4.5 G&\textbf{FA-cResNet}\\
			\hline
			\multirow{2}{*}{\textbf{HyperMS-SSIM-4}}&\multirow{2}{*}{484 ms}&289 ms&2.2 ms&177.8 M&\textbf{23.6 M}&\textbf{4.1 G}&\textbf{Anchor 2}\\
			&&-&\textbf{1.8 ms}&\textbf{110.6 M}&42.0 M&4.9 G&\textbf{FA-cResNet}\\
			\hline
			\multirow{2}{*}{\textbf{HyperMS-SSIM-8}}&\multirow{2}{*}{600 ms}&355 ms&\textbf{2.3 ms}&177.8 M&\textbf{23.6 M}&\textbf{4.1 G}&\textbf{Anchor 2}\\
			&&-&2.4 ms&\textbf{110.6 M}&42.1 M&4.9 G&\textbf{FA-cResNet}\\
			\hline
			\toprule[2pt]
	\end{tabular}}
\end{table}

Table~\ref{tab:eff} provides the computational costs per image in terms of inference time (inf. time), the total feature size during forward/backward pass (pass size), the number of parameters (\#params) and floating point operations (FLOPs) averaged over the MINC-2500 test dataset. Although our FA-cResNet requires more parameters and FLOPs, it can produce a smaller feature size and a similar inference speed as compared to Anchor 2. Actually, our FA-cResNet networks are trained and tested with compressed-domain representations without decoding. It can be seen that the decoding time for compression can be significantly slower (more than $100 \times$) than the inference time of the recognition model; thus our FA-cResNet can be much more efficient by avoiding the decoding process. The encoding process can be one-pass and it is no longer used for recognition after the original images are encoded and stored as compressed-domain representations.

\subsection{Ablation Study}

To better understand our proposed model and reveal the effectiveness of the adopted components, we conduct extensive ablation experiments on input, FA module on ResNet/cResNet, CAU/FEU, FA module position, CAU pooling function, CAU position, CAU module, CAU bottleneck architecture and FEU architecture using the MINC-2500 dataset, and provide the detailed results in Table~\ref{tab:abl}. In most ablation cases, our FA-cResNet can produce better performance results in terms of Top-1 classification accuracy (shown as bold numbers) as compared to the pixel-domain Anchor 2 despite the fact that it aims to improve the performance on compressed-domain image recognition. This shows that our FA-cResNet can still be very effective in various parameter/module settings.

\begin{table}[!tb]
	\centering
	\caption{Ablation experimental results showing obtained Top-1 classification accuracy under different settings using the MINC-2500 dataset. Underline numbers indicate best performances within each experiment and bold numbers represent that the accuracy results within the current setting of our FA-cResNet are higher (equal) as compared to Anchor 2. * denotes the default option in our FA-cResNet.}
	\label{tab:abl}
	\vspace{5pt}
	\renewcommand{\arraystretch}{1.2}
	\resizebox{1\textwidth}{!}{
		\begin{tabular}{cc}
			\bottomrule[2pt]
			&\multirow{5}{*}{
				\setlength{\tabcolsep}{2.918mm}{
					\begin{tabular}{|cc|cc|ccc}
						\multicolumn{2}{|c}{\textbf{Input}}&\multicolumn{2}{|c}{\textbf{FA Effect}}&\multicolumn{3}{|c}{\textbf{FA Module}}\\
						$\mathbf{\hat{y}}$&$\mathbf{\hat{y}\&\hat{\sigma}}\textbf{*}$&\textbf{FA-ResNet}&\textbf{FA-cResNet*}&\textbf{CAU}&\textbf{FEU}&\textbf{FA*}\\
						\hline
						\textbf{72.4}&\textbf{\textbf{73.1}}&\textbf{\textbf{73.0}}&\underline{\textbf{73.1}}&\textbf{71.8}&\textbf{72.9}&\textbf{\textbf{73.1}}\\
						\textbf{75.5}&\textbf{\textbf{76.5}}&\underline{\textbf{75.7}}&\underline{\textbf{76.5}}&74.5&\textbf{75.9}&\underline{\textbf{76.5}}\\
						\textbf{76.7}&\underline{\textbf{77.9}}&\textbf{75.4}&\underline{\textbf{77.9}}&\textbf{76.0}&\textbf{76.7}&\textbf{\textbf{77.9}}\\
			\end{tabular}}}\\
			&\\
			\hline
			\textbf{HyperMS-SSIM-1}&\\
			\textbf{HyperMS-SSIM-4}&\\
			\textbf{HyperMS-SSIM-8}&\\
			\hline
			&\multirow{5}{*}{
				\setlength{\tabcolsep}{4.375mm}{
					\begin{tabular}{|cc|ccccc}
						\multicolumn{2}{|c}{\textbf{FA Position}}&\multicolumn{5}{|c}{\textbf{CAU Pooling Function}}\\
						\textbf{pre*}&\textbf{post}&\textbf{max}&\textbf{avg}&\textbf{max\&avg}&\textbf{1att}&\textbf{2att*}\\
						\hline
						\underline{\textbf{73.1}}&\textbf{71.7}&\textbf{72.9}&\textbf{72.7}&\textbf{72.6}&\textbf{72.8}&\underline{\textbf{73.1}}\\
						\underline{\textbf{76.5}}&\textbf{75.5}&\textbf{75.8}&\textbf{76.1}&\textbf{76.1}&\textbf{75.5}&\underline{\textbf{76.5}}\\
						\underline{\textbf{77.9}}&\textbf{76.7}&\textbf{77.0}&\textbf{77.4}&\textbf{77.3}&\textbf{77.5}&\underline{\textbf{77.9}}\\
			\end{tabular}}}\\
			&\\
			\hline
			\textbf{HyperMS-SSIM-1}&\\
			\textbf{HyperMS-SSIM-4}&\\
			\textbf{HyperMS-SSIM-8}&\\
			\hline
			&\multirow{5}{*}{
				\setlength{\tabcolsep}{4.598mm}{
					\begin{tabular}{|cccc|ccc}
						\multicolumn{4}{|c}{\textbf{CAU Position}}&\multicolumn{3}{|c}{\textbf{CAU Module}}\\
						\textbf{pre*}&\textbf{skip}&\textbf{inner}&\textbf{post}&\textbf{scale}&\textbf{bias}&\textbf{affine*}\\
						\textbf{73.1}&\underline{\textbf{73.6}}&\textbf{71.8}&\textbf{72.2}&\underline{\textbf{73.1}}&\textbf{72.5}&\underline{\textbf{73.1}}\\
						\underline{\textbf{76.5}}&\textbf{76.0}&74.9&75.0&\textbf{75.4}&74.9&\underline{\textbf{76.5}}\\
						\underline{\textbf{77.9}}&\textbf{77.4}&\textbf{75.9}&\textbf{76.3}&\textbf{77.3}&\textbf{76.5}&\underline{\textbf{77.9}}\\
			\end{tabular}}}\\
			&\\
			\hline
			\textbf{HyperMS-SSIM-1}&\\
			\textbf{HyperMS-SSIM-4}&\\
			\textbf{HyperMS-SSIM-8}&\\
			\hline
			&\multirow{5}{*}{
				\setlength{\tabcolsep}{4.08mm}{
					\begin{tabular}{|cccccccc}
						\multicolumn{8}{|c}{\textbf{CAU Bottleneck Architecture}}\\
						\textbf{1fc}&\textbf{r1}&\textbf{r2}&\textbf{r4}&\textbf{r8}&\textbf{r16*}&\textbf{r32}&\textbf{r64}\\
						\textbf{72.1}&68.5&\textbf{72.0}&\textbf{72.8}&\textbf{\textbf{73.1}}&\underline{\textbf{73.1}}&\textbf{72.5}&\textbf{73.0}\\
						74.5&74.8&\textbf{75.7}&74.7&\textbf{76.1}&\underline{\textbf{76.5}}&\underline{\textbf{76.5}}&\textbf{75.4}\\
						73.0&\textbf{76.3}&\textbf{76.8}&\textbf{76.8}&\textbf{76.2}&\underline{\textbf{77.9}}&\textbf{77.3}&\textbf{77.3}\\
			\end{tabular}}}\\
			&\\
			\hline
			\textbf{HyperMS-SSIM-1}&\\
			\textbf{HyperMS-SSIM-4}&\\
			\textbf{HyperMS-SSIM-8}&\\
			\hline
			&\multirow{5}{*}{
				\setlength{\tabcolsep}{3.5mm}{
					\begin{tabular}{|cccccccc}
						\multicolumn{8}{|c}{\textbf{FEU Architecture}}\\
						\textbf{2/2/2}&\textbf{1/2/3}&\textbf{3/3/3}&\textbf{2/3/4}&\textbf{4/4/4}&\textbf{3/4/5}&\textbf{5/5/5}&\textbf{4/5/6}\\
						\textbf{72.8}&\textbf{72.3}&\textbf{72.8}&\underline{\textbf{73.1*}}&\textbf{72.6}&\textbf{72.6}&\textbf{72.5}&\textbf{73.0}\\
						\textbf{75.3}&\textbf{75.8}&\textbf{75.4}&\textbf{75.6}&\textbf{76.3}&\underline{\textbf{76.5*}}&\textbf{75.8}&\textbf{76.1}\\
						\textbf{77.1}&\textbf{77.4}&\textbf{77.1}&\textbf{77.6}&\textbf{77.0}&\underline{\textbf{77.9*}}&\textbf{76.6}&\textbf{77.0}\\
			\end{tabular}}}\\
			&\\
			\hline
			\textbf{HyperMS-SSIM-1}&\\
			\textbf{HyperMS-SSIM-4}&\\
			\textbf{HyperMS-SSIM-8}&\\
			\toprule[2pt]
	\end{tabular}}
\end{table}

\textbf{Input.} We retrain and test our FA-cResNet with only the compressed-domain representations $\hat{y}$ as input and it shows that the $\hat{y}$-only FA-cResNet provides a better result than the pixel-domain Anchor 2 and that the standard deviation maps $\hat{\sigma}$ that help in the hyperprior compression network can also further improve the compressed-domain image recognition performance.

\textbf{FA Effect.} We also insert the FA modules into Layers 1-4 of the baseline ResNet and retrain the resulting network from scratch with the decoded images. The obtained results indicate that the proposed FA-cResNet can produce clearly better Top-1 accuracy results under HyperMS-SSIM-4/HyperMS-SSIM-8 than FA-ResNet, which can imply that the proposed FA module is more suitable for improving the cResNet in the compressed-domain as compared to improving the pixel-domain ResNet.

\textbf{FA Module.} While our model can already produce a similar or better Top-1 recognition accuracy with either CAU-only or FEU-only as compared to Anchor 2, these two modules can be complementary and result in the best results when combined as the FA module.

\textbf{FA Position.} We compare the effect of FA module position by imposing our FA modules after the first blocks (pre) / the last blocks (post) of Layers 2 - 4. And it shows that inserting the FA module in a layer performs better and is preferred for feature adaptation as compared to having the FA module after the last block for a layer.

\textbf{CAU Pooling Function.} We test different pooling functions to extract the channel-wise attention vectors. The nonparametric pooling models (max, avg, max\&avg) give similar recognition results, which can be further improved by the attention pooling function. Also, as compared to shared attention pooling (1att) for channel-wise weight and bias, independent attention poolings (2att) yields a better result.

\textbf{CAU Position.} By inserting CAU before the skip branch (pre), in the skip connection (skip), right before the residual addition (inner) and after the residual addition (post) of FEU, we find that adapting the feature maps with the channel information by using CAU before FEU is preferred.

\textbf{CAU Module.} We also run experiments on either channel-wise scaling ($\mathbf{\Tilde{x}} = \mathbf{w} \cdot \mathbf{x}$) or channel-wise bias ($\mathbf{\Tilde{x}} = \mathbf{x} + \mathbf{b}$) in CAU, showing that the channel-wise affine transform ($\mathbf{\Tilde{x}} = \mathbf{w} \cdot \mathbf{x} + \mathbf{b}$) is necessary to obtain the best performance.

\textbf{CAU Bottleneck Architecture.} It can be seen that the two-linear-layer bottleneck architecture with a relatively large reduction ratio $r$ generally outperforms the one with a small reduction ratio $r$ and the one-linear-layer architecture, which implies the necessity of reducing the number of channels and then recovering it in the two-linear-layer module in the compressed-domain.

\textbf{FEU Architecture.} We modify the number of $3 \times 3$ convolutional blocks $n$ in FEU and also test the cases of identical and increasing $n$ in the FA modules of Layers 2-4. It can be observed that an increasing $n$ for different stages can generally produce a better result as compared to a fixed $n$.

\subsection{Effects of Compression Quality Metrics}

\begin{figure*}[!t]
	\centering
	\begin{subfigure}{0.32\textwidth}
		\centering
		\includegraphics[width=1\textwidth]{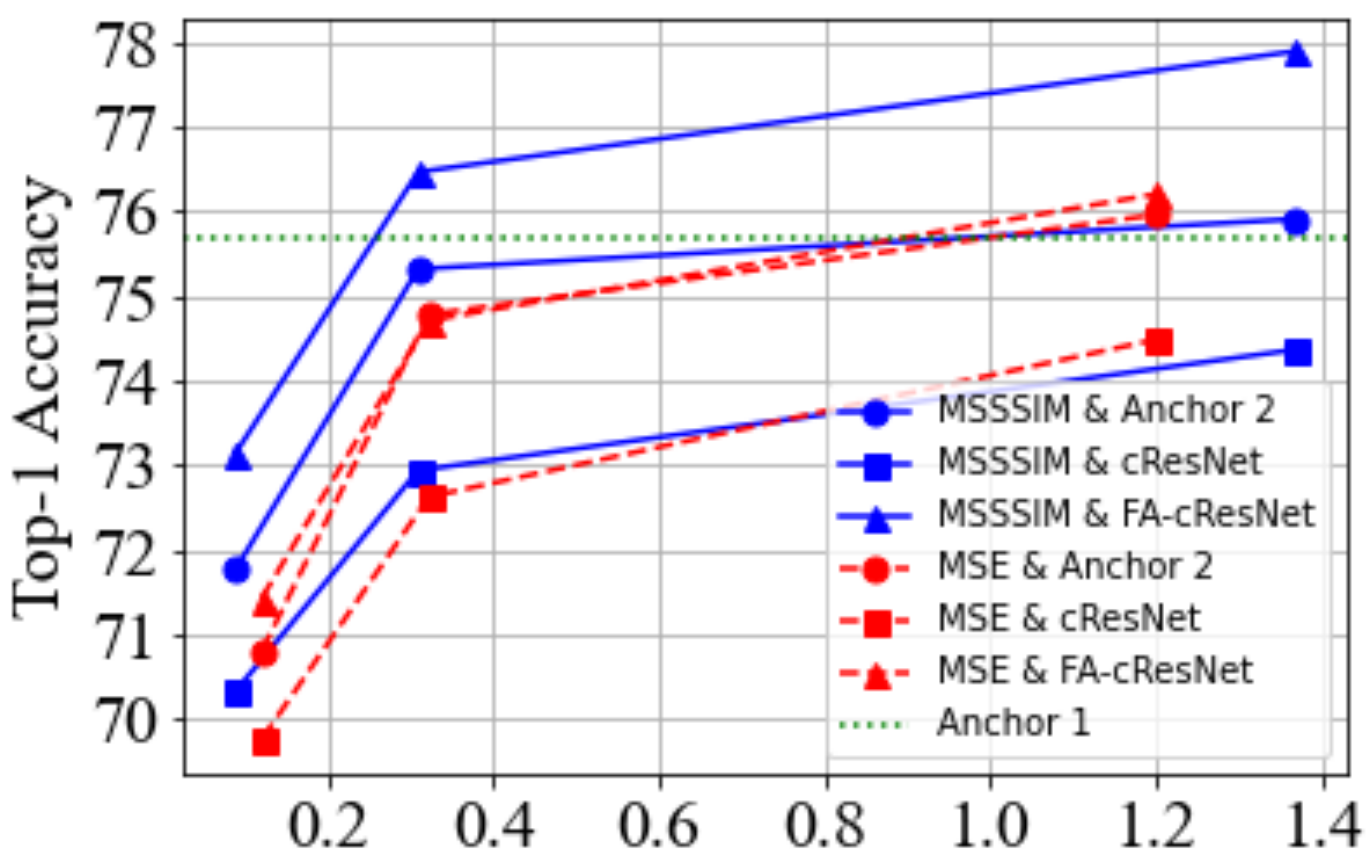}
		\caption{Bit Rate}
		\label{fig:bit}
	\end{subfigure}
	\begin{subfigure}{0.32\textwidth}
		\centering
		\includegraphics[width=1\textwidth]{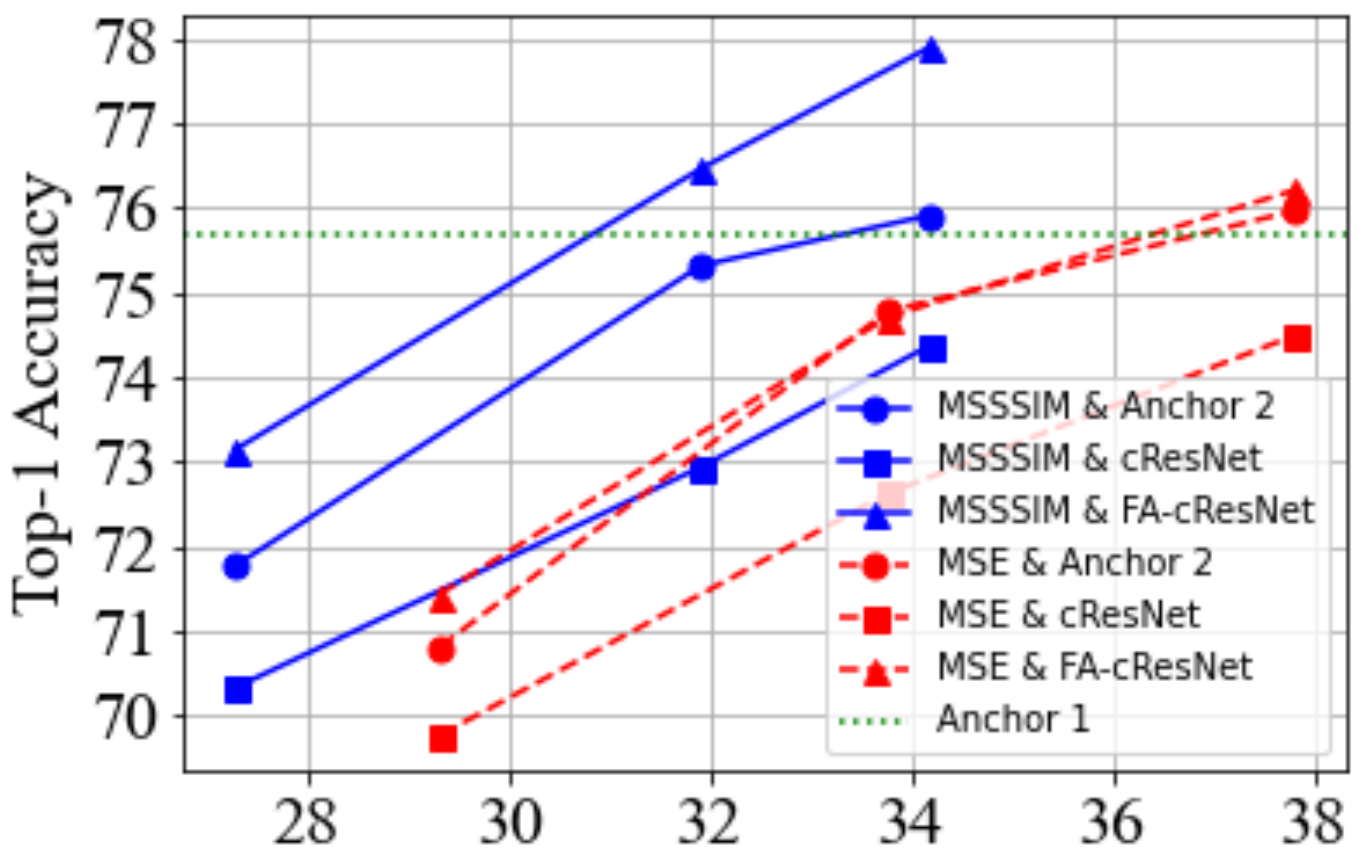}
		\caption{PSNR}
		\label{fig:psnr}
	\end{subfigure}
	\begin{subfigure}{0.32\textwidth}
		\centering
		\includegraphics[width=1\textwidth]{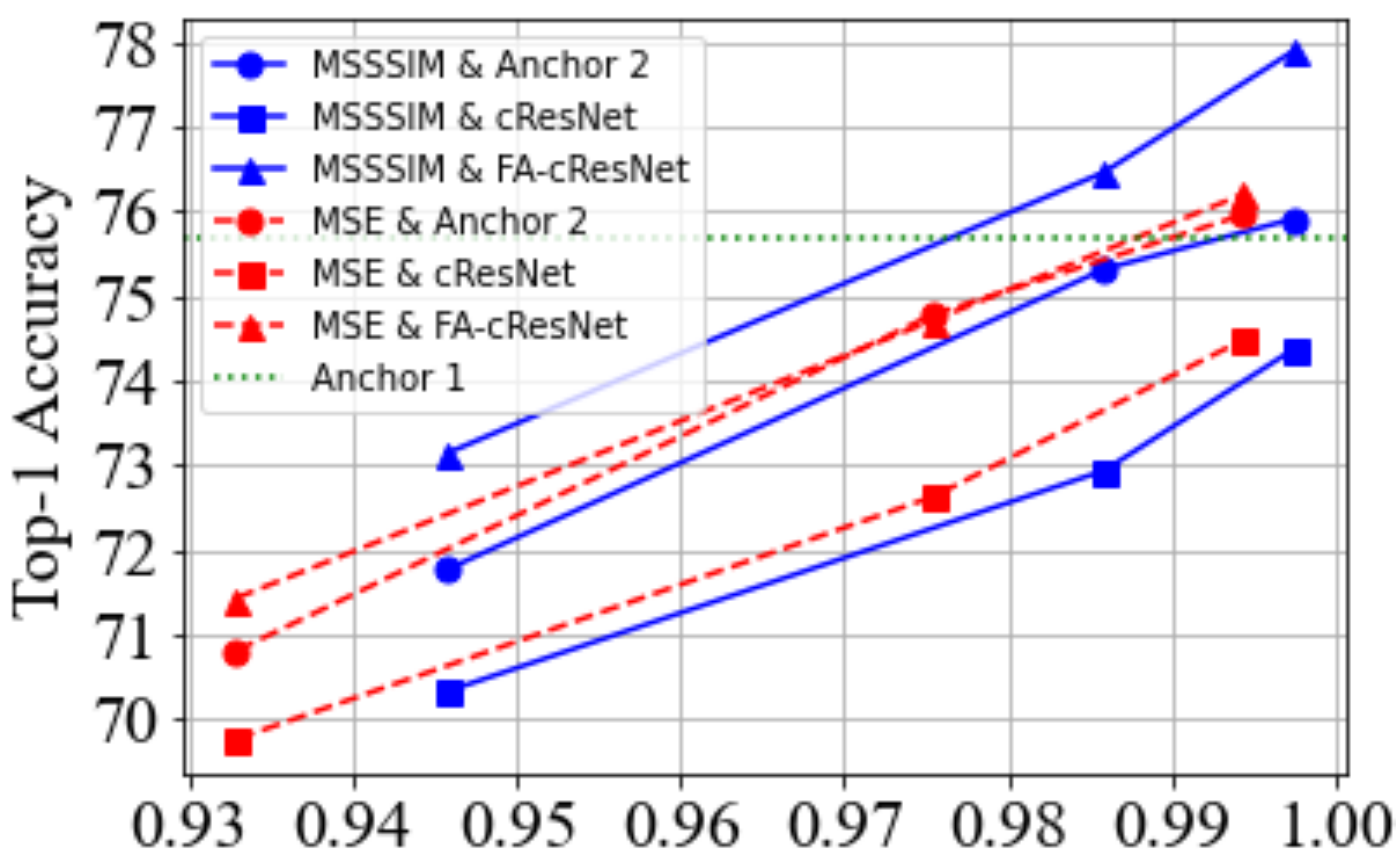}
		\caption{MS-SSIM}
		\label{fig:msssim}
	\end{subfigure}
	\caption{Curves for Top-1 accuracy results vs. compression rate/qualities (bit rate, PSNR and MS-SSIM) when using hte HyperMS-SSIM and HyperMSE compression models.}
	\label{fig:qual}
\end{figure*}

In general, MS-SSIM~\cite{wang2003multiscale}/SSIM~\cite{wang2004image} can be closer to the subjective image quality assessment by human vision as compared to PSNR. However, it is unclear which metric is more suitable for machine vision. Thus, we also train and examine the texture recognition networks with pixel-domain decoded images and compressed-domain representations under the same compression network that is pretrained with the MSE quality metric in the loss function (HyperMSE~\cite{balle2018variational}). Figure~\ref{fig:qual} provides the accuracy curves of pixel-domain classifications (Anchors 1\&2) and compressed-domain classifications (cResNet \& FA-cResNet) using HyperMS-SSIM and HyperMSE compression models separately, and shows their trends for different compression metrics (bit rate, PSNR, MS-SSIM) that are computed on the MINC-2500 test split 1.

As shown in Figure~\ref{fig:bit}, the accuracy results increase as the bit rate grows. Compared to the pixel-domain and compressed-domain image recognition models with inputs that are compressed by HyperMS-SSIM, using inputs compressed by HyperMSE produces lower accuracy results, except for slightly better performance in terms of Top-1 accuracy using the Anchor 2 and cResNet models with HyperMSE-8. It can be seen that our FA-cResNet can still produce a similar or better performance when compared with Anchor 2 for each bit rate of HyperMSE, while cResNet lags behind with a clear performance gap with the other two models. Generally, Figures~\ref{fig:psnr}\&\ref{fig:msssim} show that HyperMS-SSIM can yield a better MS-SSIM output while HyperMSE provides a higher PSNR result for their compression models with the same index number (1/4/8) under a similar compression bit rate, which corresponds to their respective quality metrics in the compression loss function. In Figure~\ref{fig:psnr}, the classification models under HyperMS-SSIM give a generally better performance trade-off between accuracy and PSNR than those under HyperMSE; in Figure~\ref{fig:msssim}, the accuracy curves of Anchor 2 under HyperMS-SSIM and HyperMSE show a nearly linear proportion between accuracy and MS-SSIM. Overall, our FA-cResNet trained with the outputs of HyperMS-SSIM can produce the best accuracy results regardless of the considered compression metric.

\section{Conclusion}
\label{sec:con}

We propose a learning-based compressed-domain image recognition framework that integrates a novel feature adaptation module. The proposed feature adaptation module consists of a channel-wise attention unit (CAU) and feature enhancement unit (FEU). Specifically, the CAU aims to learn the channel-wise affine transform vectors for selecting and realigning compressed-domain channels, while the FEU aims to enhance the useful features within the selected channels by means of learned cross-channel convolutional layers. We perform both natural image recognition and texture image recognition tasks on the ImageNet and MINC-2500 datasets, respectively, with HyperMS-SSIM learning-based compression network and compare the obatined classification accuracy results with pixel-domain anchor models as well as existing compressed-domain ones. The experimental results show that our FA-cResNet can outperform the existing compressed-domain methods without decoding under similar compression rates on ImageNet, and it can also improve the accuracy results of cResNet to outperform the pixel-domain decoded image trained models (Anchor 2). By avoiding the decoding process, our FA-cResNet can significantly reduce the computational time as compared to Anchor 2. An extensive ablation study shows that the proposed architecture can still result in better results than Anchor 2 in various settings on parameters/modules. By illustrating the results versus different compression quality measurements, we demonstrate that the HyperMS-SSIM learning-based compression model can be more favorable for our proposed FA-cResNet recognition model as compared to the HyperMSE model.

{\small
\bibliographystyle{ieee_fullname}
\bibliography{egbib}
}

\end{document}